\PassOptionsToPackage{unicode}{hyperref}
\PassOptionsToPackage{hyphens}{url}
\PassOptionsToPackage{dvipsnames,svgnames,x11names}{xcolor}
\documentclass[
  12pt,
]{article}
\usepackage{xcolor}
\usepackage{mathptmx}
\usepackage[margin=1in]{geometry}
\usepackage{amsmath,amssymb}
\usepackage{iftex}
\ifPDFTeX
\usepackage{newtxtext, newtxmath}
  \usepackage[utf8]{inputenc}
  \usepackage{textcomp} 
\else 
  \usepackage{unicode-math} 
  \defaultfontfeatures{Scale=MatchLowercase}
  \defaultfontfeatures[\rmfamily]{Ligatures=TeX,Scale=1}
\fi
\usepackage{lmodern}

\IfFileExists{upquote.sty}{\usepackage{upquote}}{}
\IfFileExists{microtype.sty}{
  \usepackage[]{microtype}
  \UseMicrotypeSet[protrusion]{basicmath} 
}{}
\usepackage{setspace}
\makeatletter
\@ifundefined{KOMAClassName}{
  \IfFileExists{parskip.sty}{%
    \usepackage{parskip}
  }{
    \setlength{\parindent}{0pt}
    \setlength{\parskip}{6pt plus 2pt minus 1pt}}
}{
  \KOMAoptions{parskip=half}}
\makeatother
\makeatletter
\ifx\paragraph\undefined\else
  \let\oldparagraph\paragraph
  \renewcommand{\paragraph}{
    \@ifstar
      \xxxParagraphStar
      \xxxParagraphNoStar
  }
  \newcommand{\xxxParagraphStar}[1]{\oldparagraph*{#1}\mbox{}}
  \newcommand{\xxxParagraphNoStar}[1]{\oldparagraph{#1}\mbox{}}
\fi
\ifx\subparagraph\undefined\else
  \let\oldsubparagraph\subparagraph
  \renewcommand{\subparagraph}{
    \@ifstar
      \xxxSubParagraphStar
      \xxxSubParagraphNoStar
  }
  \newcommand{\xxxSubParagraphStar}[1]{\oldsubparagraph*{#1}\mbox{}}
  \newcommand{\xxxSubParagraphNoStar}[1]{\oldsubparagraph{#1}\mbox{}}
\fi
\makeatother

\usepackage{longtable,booktabs,array}
\usepackage{calc} 
\usepackage{etoolbox}
\makeatletter
\patchcmd\longtable{\par}{\if@noskipsec\mbox{}\fi\par}{}{}
\makeatother
\IfFileExists{footnotehyper.sty}{\usepackage{footnotehyper}}{\usepackage{footnote}}
\makesavenoteenv{longtable}
\usepackage{graphicx}
\makeatletter
\newsavebox\pandoc@box
\newcommand*\pandocbounded[1]{
  \sbox\pandoc@box{#1}%
  \Gscale@div\@tempa{\textheight}{\dimexpr\ht\pandoc@box+\dp\pandoc@box\relax}%
  \Gscale@div\@tempb{\linewidth}{\wd\pandoc@box}%
  \ifdim\@tempb\p@<\@tempa\p@\let\@tempa\@tempb\fi
  \ifdim\@tempa\p@<\p@\scalebox{\@tempa}{\usebox\pandoc@box}%
  \else\usebox{\pandoc@box}%
  \fi%
}
\def\fps@figure{htbp}
\makeatother

\NewDocumentCommand\citeproctext{}{}

\makeatletter
 \let\@cite@ofmt\@firstofone
 \def\@biblabel#1{}
 \def\@cite#1#2{{#1\if@tempswa , #2\fi}}
\makeatother
\newlength{\cslhangindent}
\newlength{\csllabelwidth}
\newenvironment{CSLReferences}[2] 
 {\begin{list}{}{%
  \setlength{\itemindent}{0pt}
  \setlength{\leftmargin}{0pt}
  \setlength{\parsep}{0pt}
  \ifodd #1
   \setlength{\leftmargin}{\cslhangindent}
   \setlength{\itemindent}{-1\cslhangindent}
  \fi
  \setlength{\itemsep}{#2\baselineskip}}}
 {\end{list}}
\usepackage{calc}

\providecommand{\tightlist}{%
  \setlength{\itemsep}{0pt}\setlength{\parskip}{0pt}}

\usepackage{booktabs}
\usepackage{longtable}
\usepackage{array}
\usepackage{multirow}
\usepackage{wrapfig}
\usepackage{float}
\usepackage{colortbl}
\usepackage{pdflscape}
\usepackage{tabu}
\usepackage{threeparttable}
\usepackage{threeparttablex}
\usepackage[normalem]{ulem}
\usepackage{makecell}
\usepackage{xcolor}
\usepackage{setspace}
\makeatletter
\def\keywords#1{\vspace{0.5em}\noindent\textbf{Keywords:} #1\par}
\makeatother
\usepackage{etoolbox}
\AtBeginEnvironment{tabular}{\singlespacing}
\AtBeginEnvironment{longtable}{\singlespacing}
\makeatletter
\@ifpackageloaded{caption}{}{\usepackage{caption}}
\AtBeginDocument{%
\ifdefined\contentsname
  \renewcommand*\contentsname{Table of contents}
\else
  \newcommand\contentsname{Table of contents}
\fi
\ifdefined\listfigurename
  \renewcommand*\listfigurename{List of Figures}
\else
  \newcommand\listfigurename{List of Figures}
\fi
\ifdefined\listtablename
  \renewcommand*\listtablename{List of Tables}
\else
  \newcommand\listtablename{List of Tables}
\fi
\ifdefined\figurename
  \renewcommand*\figurename{Figure}
\else
  \newcommand\figurename{Figure}
\fi
\ifdefined\tablename
  \renewcommand*\tablename{Table}
\else
  \newcommand\tablename{Table}
\fi
}
\@ifpackageloaded{float}{}{\usepackage{float}}
\floatstyle{ruled}
\@ifundefined{c@chapter}{\newfloat{codelisting}{h}{lop}}{\newfloat{codelisting}{h}{lop}[chapter]}
\floatname{codelisting}{Listing}

\makeatother
\usepackage{mathrsfs}
\makeatletter
\@ifpackageloaded{caption}{}{\usepackage{caption}}
\@ifpackageloaded{subcaption}{}{\usepackage{subcaption}}
\makeatother
\usepackage{bookmark}
\IfFileExists{xurl.sty}{\usepackage{xurl}}{} 
\hypersetup{
  pdftitle={An Enhanced Projection Pursuit Tree Classifier with Visual Methods for Assessing Algorithmic Improvements},
  pdfkeywords={classification, heterogeneous variance, machine
learning, projection pursuit, separability, tree classifier},
  colorlinks=true,
  linkcolor={blue},
  filecolor={Maroon},
  citecolor={Blue},
  urlcolor={Blue},
  pdfcreator={LaTeX via pandoc}}

\title{An Enhanced Projection Pursuit Tree Classifier with Visual
Methods for Assessing Algorithmic Improvements}
\author{}
\date{}
\begin{document}
\maketitle

\setstretch{2}
\author{}
\date{}
\maketitle

\begin{center}
{\large Natalia da Silva$^{1}$ \quad Dianne Cook$^{2}$ \quad Eun-Kyung Lee$^{3}$}

\vspace{0.3em}
{\small
$^{1}$Instituto de Estadística (IESTA), Universidad de la República, Montevideo, Uruguay\\
$^{2}$Department of Econometrics and Business Statistics, Monash University, Melbourne, Australia\\
$^{3}$Department of Statistics, Ewha Womans University, Seoul, Republic of Korea
}
\end{center}

\vspace{1em}

\begin{abstract}
This paper presents enhancements to the projection pursuit tree classifier and visual diagnostic methods for assessing their impact in high dimensions. The original algorithm uses linear combinations of variables in a tree structure where depth is constrained to be less than the number of classes---a limitation that proves too rigid for complex classification problems. Our extensions improve performance in multi-class settings with unequal variance-covariance structures and nonlinear class separations by allowing more splits and more flexible class groupings in the projection pursuit computation. Proposing algorithmic improvements is straightforward; demonstrating their actual utility is not. We therefore develop two visual diagnostic approaches to verify that the enhancements perform as intended. Using high-dimensional visualization techniques, we examine model fits on benchmark datasets to assess whether the algorithm behaves as theorized. An interactive web application enables users to explore the behavior of both the original and enhanced classifiers under controlled scenarios. The enhancements are implemented in the R package \texttt{PPtreeExt}, which is available on CRAN.
\end{abstract}


\section{Introduction}\label{sec-introduction}

Loh (2014) provides an extensive overview of decades of research on
classification and regression tree algorithms. Tree models are useful
because they are simple, yet flexible, fast to compute, widely
applicable, and interpretable. A main drawback, though, is that most
available methods use splits along single variables, making separations
on a combination of variables harder to capture. The reason is that it
is a much harder problem to optimize the search for linear combinations,
and thus, there is a lack of useful software that provides oblique
splits.

One available method and software is the projection pursuit tree
(PPtree) (Lee et al. 2013). It optimizes a projection pursuit (PP)
index, for example, the LDA (Lee et al. 2005) or PDA (Lee and Cook 2010)
index among others, based on class information to identify
one-dimensional projections that best separate the groups at each node.
The PPtree tree structure is simpler than classic methods like
\texttt{rpart} (Therneau and Atkinson 2025), as it restricts the tree
depth to \(G-1\) (\(G\) is the number of classes). At each node, PPtree
separates the data into two groups by first deciding which classes to
group together; the data are split into left and right branches based on
two group means. This produces a compact and interpretable tree
implemented in the \texttt{PPtreeViz} R package (Lee 2018).

One of the main advantages of PPtree is its ability to leverage the
correlation between predictor variables to improve class separation. In
some scenarios, it has been shown to outperform other classifiers,
including random forests (RF) (Breiman 2001). PP addresses a known
limitation of the original RF algorithm. It included oblique
projections, but because it is based on randomly generated splits was
practically infeasible. It is an inefficient algorithm for finding
useful oblique splits.

The compact nature of PPtree means that it fits simply and does not
require pruning. The resulting one-dimensional projections can be used
to make visualizations of group separations. However, there are
limitations that reduce its effectiveness. In multiclass problems, due
to the way classes are grouped, the prediction boundaries are often too
close to one group, resulting in unnecessarily high error rates. This
happens because grouping of classes likely produces unequal
variance-covariances, making the pooled variance-covariance used by LDA
inadequate. Secondly, when classes have non-elliptical
variance-covariance, or have multiple clusters per class, the rigidity
of the \((G\)-1)-node structure is not sufficiently flexible. The
algorithm is modified to address these limitations in two ways: (1)
improving prediction boundaries by modifying the choice of split points
through class subsetting; and (2) increasing flexibility by allowing
multiple splits per group.

Proposing enhancements to algorithms is easy, and theoretically, the
modifications may seem sensible. In practice, they may not work as one
envisions. Thus, an important part of developing algorithm modifications
is providing validation that they are doing what is expected, or not.
This means going beyond predictive accuracy to understanding the
resulting models. A second objective of this paper is to provide visual
diagnostics for the algorithm changes. This is two-fold: a shiny (Chang
et al. 2025) app and tours to view the fits in high dimensions. The app
is available in the \texttt{PPtreeExt} package (da Silva et al. 2026)
along with the implementation of the modified algorithm. It allows users
to generate 2D data sets with different types of separations and compare
the model fits for the modified and the original algorithm. This was an
essential tool in constructing the new algorithm, because it provides a
side-by-side visualization of prediction boundaries to identify data
patterns in which the different methods perform better. The second
visual diagnostic is to use tours available in the \texttt{tourr}
package (Wickham et al. 2011) to examine the high-dimensional structure
of data sets and the resulting model fits, where the modified algorithm
outperforms other classifiers. This explains the performance comparison
results and intuition for why the modified algorithm performs better.

The paper is organized as follows: Section~\ref{sec-pptal} provides
background on the original PPtree algorithm, including its structure and
limitations. Section~\ref{sec-ppext} describes the extensions to the
algorithm. Section~\ref{sec-compare} contains the examples on visually
and numerically digesting the enhanced algorithm's performance. Finally,
Section~\ref{sec-disc} discusses the changes, results of the comparison,
and potential future directions.

\section{Background to PPtree}\label{sec-pptal}

\begin{figure}

\centering{

\includegraphics[width=1\linewidth,height=\textheight,keepaspectratio]{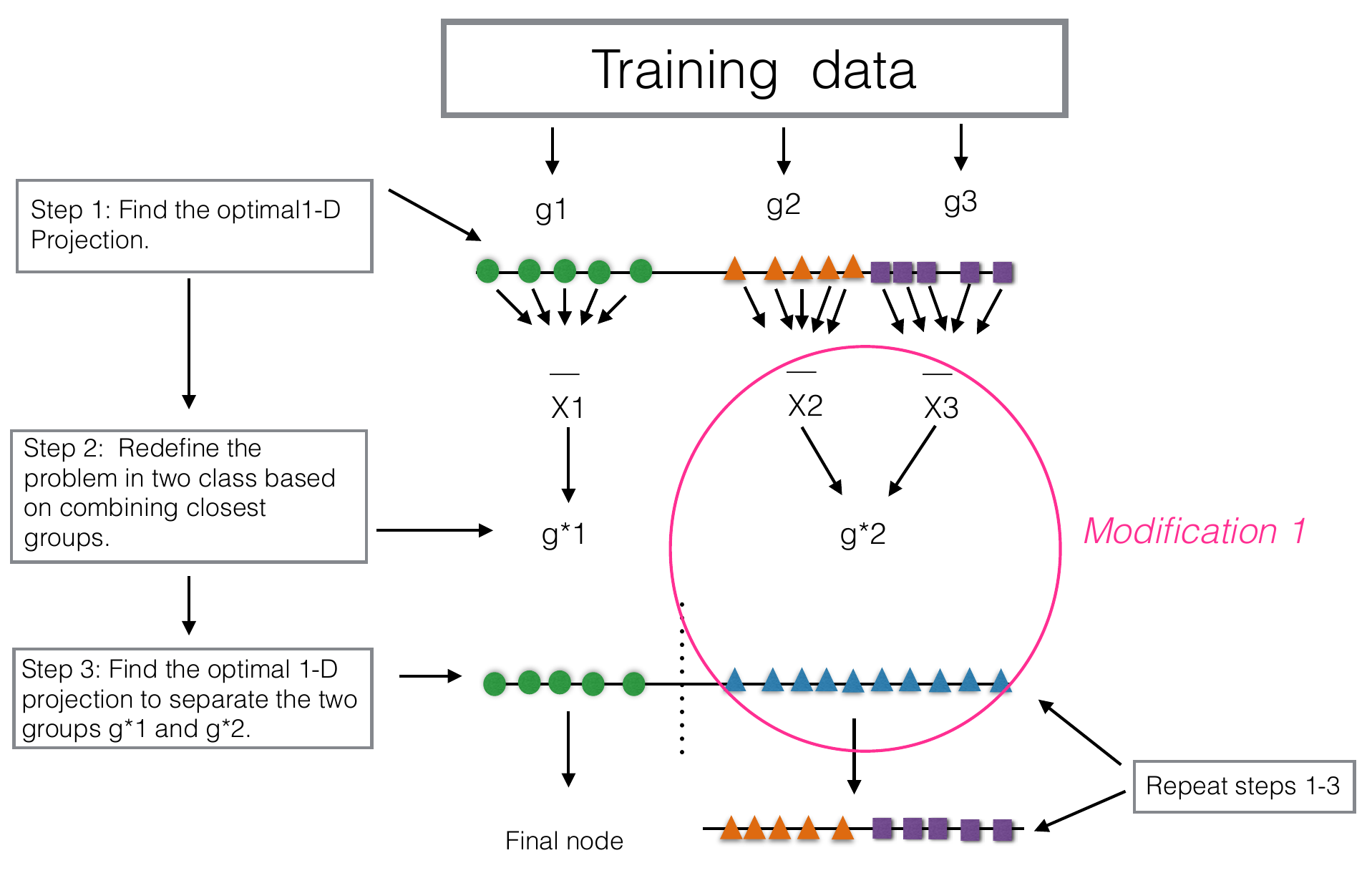}

}

\caption{\label{fig-diag}Illustration of the original PPtree algorithm
for \(G=3\). In step 1, PP optimization decides on the best linear
projection. Step 2 combines two groups to make a two-group
classification. The PP optimization is computed for two group problem,
and an appropriate split is decided. This is repeated for the remaining
two groups to get the final model containing two splits. The locations
where modifications will be made are circled.}

\end{figure}

Figure~\ref{fig-diag} illustrates the original PPtree algorithm for
\(G=3\). Starting from the optimal projection, the left-side, most
distant, cluster (green circles) is relabeled as \(g_1^*\), while the
other two clusters are combined and relabeled as \(g_2^*\). A second
projection vector is then computed using all observations, following the
same procedure applied to these two new super-groups. A split point is
determined using one of the predefined rules, and observations are
assigned to branches by comparing their projected values to the cutoff.

The algorithm computes a split point using one of several methods, and
the observations are allocated to groups based on whether their
projected values fall below or above the cutoff. Since \(g_1^*\)
contains only one of the original classes, the left branch becomes a
terminal node. Because \(g_2^*\) still includes two original classes,
the procedure is recursively repeated on the right branch.

There are eight different rules available for computing the split point
on a given projection. Let \(\bar{\mathbf{x}}_g\),
\(\mathbf{x}_g^{med}\), \(s_g\), \(IQR_g\), and \(n_g\) denote the mean,
median, standard deviation, interquartile range, and sample size of
group \(g\), respectively. Table~\ref{tbl-rules} summarizes the formulas
used to calculate the split point value, \(c\), under each rule.

\begin{longtable}[]{@{}
  >{\raggedright\arraybackslash}p{(\linewidth - 4\tabcolsep) * \real{0.1333}}
  >{\raggedright\arraybackslash}p{(\linewidth - 4\tabcolsep) * \real{0.2667}}
  >{\raggedright\arraybackslash}p{(\linewidth - 4\tabcolsep) * \real{0.6000}}@{}}
\caption{Rules for computing the split value, \(c\), between two groups,
on a data projection, using means, weighted mean, medians, standard
deviations, and interquartile ranges.}\label{tbl-rules}\tabularnewline
\toprule\noalign{}
\begin{minipage}[b]{\linewidth}\raggedright
Rule
\end{minipage} & \begin{minipage}[b]{\linewidth}\raggedright
Name
\end{minipage} & \begin{minipage}[b]{\linewidth}\raggedright
Formula
\end{minipage} \\
\midrule\noalign{}
\endfirsthead
\toprule\noalign{}
\begin{minipage}[b]{\linewidth}\raggedright
Rule
\end{minipage} & \begin{minipage}[b]{\linewidth}\raggedright
Name
\end{minipage} & \begin{minipage}[b]{\linewidth}\raggedright
Formula
\end{minipage} \\
\midrule\noalign{}
\endhead
\bottomrule\noalign{}
\endlastfoot
1 & mean &
\(c = \frac{1}{2}\bar{\mathbf{x}}_1 + \frac{1}{2}\bar{\mathbf{x}}_2\) \\
2 & sample size weighted mean &
\(c = \left(\frac{n_2}{n_1+n_2}\right)\bar{\mathbf{x}}_1 + \left(\frac{n_1}{n_1+n_2}\right)\bar{\mathbf{x}}_2\) \\
3 & standard deviation weighted mean &
\(c = \left(\frac{s_2}{s_1+s_2}\right)\bar{\mathbf{x}}_1 + \left(\frac{s_1}{s_1+s_2}\right)\bar{\mathbf{x}}_2\) \\
4 & standard error weighted mean &
\(c = \frac{s_2/\sqrt{n_2}}{s_1/\sqrt{n_1}+s_2/\sqrt{n_2}}\bar{\mathbf{x}}_1 + \frac{s_1/\sqrt{n_1}}{s_1/\sqrt{n_1}+s_2/\sqrt{n_2}}\bar{\mathbf{x}}_2\) \\
5 & median & \(c = \frac{1}{2} med_1 + \frac{1}{2} med_2\) \\
6 & sample size weighted median &
\(c = \frac{n_2}{n_1+n_2} med_1 + \frac{n_1}{n_1+n_2} med_2\) \\
7 & IQR weighted median &
\(c = \frac{IQR_2}{IQR_1+IQR_2} med_1 + \frac{IQR_1}{IQR_1+IQR_2} med_2\) \\
8 & sample size and IQR weighted median &
\(c = \frac{IQR_2/\sqrt{n_2}}{IQR_1/\sqrt{n_1} + IQR_2/\sqrt{n_2}} med_1 + \frac{IQR_1/\sqrt{n_1}}{IQR_1/\sqrt{n_1} + IQR_2/\sqrt{n_2}} med_2\) \\
\end{longtable}

\subsection{Algorithm}\label{algorithm}

The PPtree algorithm (Lee et al. 2013) follows a multi-step procedure to
identify linear combinations of predictors that optimally separate
classes in multi-class settings. Let
\(d_n = \{(\mathbf{x}i, y_i)\}_{i=1}^n\) denote the dataset, where
\(\mathbf{x}_i\) is a \(p\)-dimensional vector of explanatory variables.
The corresponding class label is \(y_i \in \mathscr{G}\), with
\(\mathscr{G} = \{1, 2, \ldots, G\}\), for \(i = 1, \ldots, n\). The
construction of a PPtree classifier follows a four-step algorithm, as
follows:

\textbf{Input:} Dataset \(d_n = \{(\mathbf{x}_i, y_i)\}_{i=1}^n\), with
\(\mathbf{x}_i \in \mathbb{R}^p\),
\(y_i \in \mathscr{G} = \{1, \ldots, G\}\)

\textbf{Output:} A projection pursuit classification tree with at most
\(G-1\) internal nodes

\begin{itemize}
\item
  \textbf{Step 1: Class Separation.} Optimize a PP index (e.g., LDA or
  PDA) to find a projection vector \(\boldsymbol{\alpha}\) that best
  separates all classes in the current node.
\item
  \textbf{Step 2: Binary Relabeling.} On projected data,
  \(z_i = \boldsymbol{\alpha}^T \mathbf{x}_i\), compute projected class
  means \(\bar{z}_1, \ldots, \bar{z}_G\). Identify the pair with the
  largest mean difference and compute the midpoint \(mp\). Relabel the
  observations into two super-classes, \(g_1^*\) and \(g_2^*\),
  depending on whether \(\bar{z}_g\) is smaller than the midpoint
  \(mp\).
\item
  \textbf{Step 3: Node Splitting.} Optimize a new projection
  \(\boldsymbol{\alpha}^*\) to separate \(g_1^*\) and \(g_2^*\) based on
  the relabeled classes. Define the split value \(c\) on
  \(z_i^* = \boldsymbol{\alpha}^{*T} \mathbf{\bar x}_{gi^*}\) using a
  predefined rule (e.g., mean or weighted mean). Split the node using
  the rule \(z_i^* < c\).
\item
  \textbf{Step 4: Recursive Partitioning.} Repeat Steps 1--3 for each
  child node until every terminal node contains only one original class.
\end{itemize}

\subsection{Less desirable aspects}\label{less-desirable-aspects}

\begin{figure}

\centering{

\pandocbounded{\includegraphics[keepaspectratio]{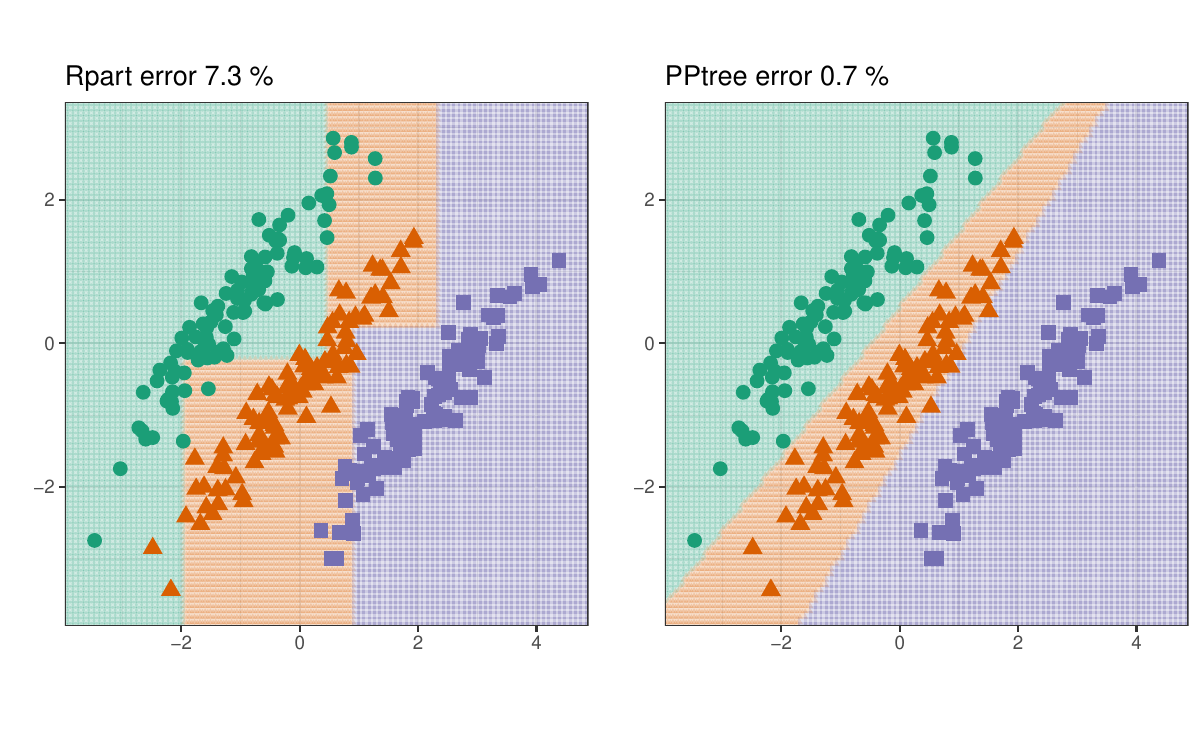}}

}

\caption{\label{fig-boundary}Comparison of decision boundaries produced
by the \texttt{rpart} (left) and PPtree (right) algorithms on the
\texttt{LC} data. The boundaries generated by PPtree are oblique to the
coordinate axes, appropriately capturing the separations as linear
combinations of the two variables. Even with many splits, the
\texttt{rpart} algorithm cannot elegantly classify this data.}

\end{figure}

\begin{figure}

\centering{

\pandocbounded{\includegraphics[keepaspectratio]{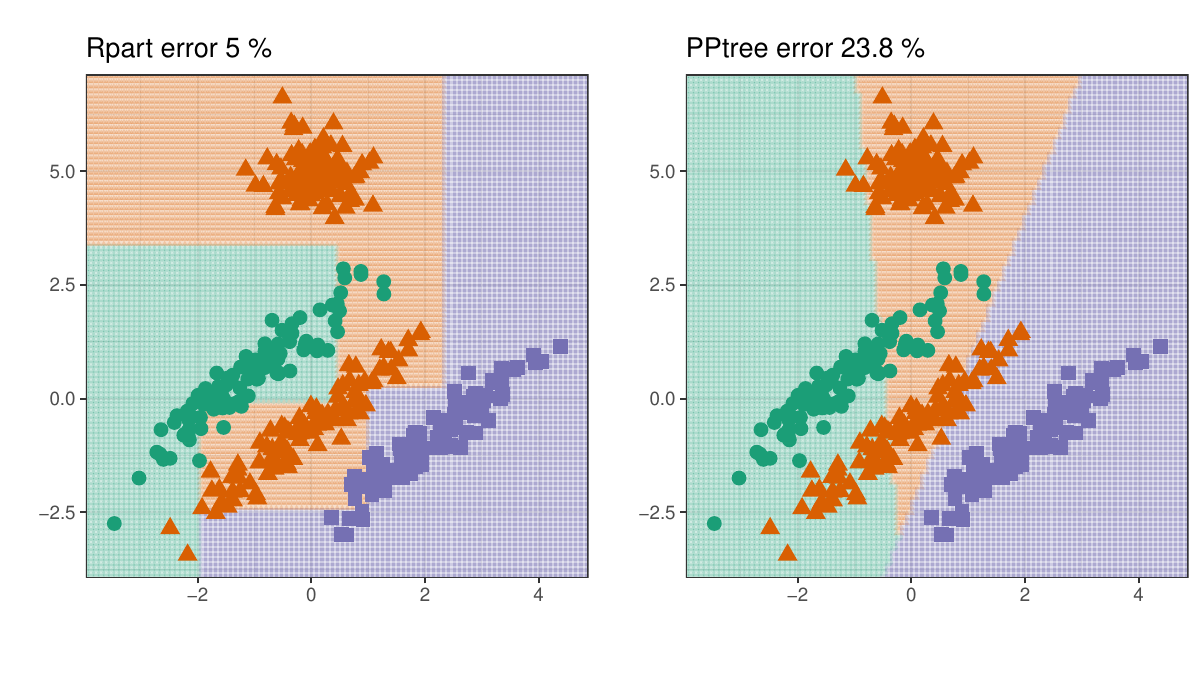}}

}

\caption{\label{fig-bound2}Comparison of decision boundaries produced by
the \texttt{rpart} (left) and PPtree (right) algorithms on the
\texttt{MC} data. The T class (orange triangles) cannot be separated
using a single linear partition, and \texttt{PPtree} fails to model it
accurately because each class must be assigned to a single terminal
node. This illustrates the need for allowing more splits in PPtree to
capture unusually shaped clusters and separations involving combinations
of variables.}

\end{figure}

The decision boundaries of a classifier reflect how the model partitions
the feature space and adapts to class separations. Visualizing the
decision boundaries provides valuable insights into model behavior and
limitations. To illustrate the PPtree algorithm relative to the
classical \texttt{rpart} and the motivation for improving PPtree, we use
two simulated three-class 2D data sets. Data set \texttt{LC} has
separations on linear combinations of variables
(Figure~\ref{fig-boundary}), and data set \texttt{MC} has multiple
clusters in a class (Figure~\ref{fig-bound2}). The three classes are
indicated by squares (S, violet), circles (C, green), and triangles (T,
orange).

Figure~\ref{fig-boundary} shows the decision boundaries produced by
\texttt{rpart} (left) and \texttt{PPtree} (right), and percentage error,
for the \texttt{LC} data. The class clusters are best separated using
linear combinations of the two variables. The \texttt{rpart} algorithm,
which relies on axis-aligned splits, must approximate the ideal boundary
using rectangular steps. Even with many steps, it cannot be as efficient
as the oblique splits provided by PPtree. The two steps of PPtree are to
split the S cluster (violet) first, and then split C (green) and T
(orange). Although the group structure is similar across the three
classes, the resulting decision boundaries are not parallel. This occurs
because the initial split uses the combined information from the C and T
clusters - treated as a super class - to compute the projection and
determine the splitting point. This also results in the boundary between
T and S being too close to T. Alternative rules for computing the split
value (see Table~\ref{tbl-rules}) can impact the position and
orientation of these decision boundaries, but in this case does not fix
the deficiency.

Figure~\ref{fig-bound2} compares the \texttt{rpart} and PPtree decision
boundaries for the data set \texttt{MC}, where class T has two separated
clusters. It highlights the limited flexibility of the decision
boundaries produced by \texttt{PPtree}. In this scenario, the T cannot
be separated using a single linear partition, and \texttt{PPtree} fails
to capture it because the tree depth is limited to a single terminal
node for each class. In contrast, the more traditional \texttt{rpart}
algorithm performs better in this case, as its recursive structure
allows it to accommodate the relatively simple non-linear separation.
However, \texttt{rpart} fails to capture the separation between classes
as linear combinations. This shows why more splits should be allowed for
the PPtree algorithm.

\section{PPtree extensions}\label{sec-ppext}

The algorithm has been extended in two main ways, as indicated by the
annotation in Figure~\ref{fig-diag}. Section~\ref{sec-algocomp} explains
how we can compare the changes these modifications induce on the
boundaries relative to \texttt{rpart} and PPtree using the Shiny app.

\subsection{Modification 1: Subsetting classes to produce better
boundaries}\label{modification-1-subsetting-classes-to-produce-better-boundaries}

The first modification targets the re-grouping step of the original
algorithm. Rather than combining all remaining classes into a single
super-class, only the two closest ones, as identified by the smallest
difference in their projected means, are used to compute the next
projection and define the split point. This step is now as follows:

\begin{itemize}
\tightlist
\item
  \textbf{Step 2\(^*\): Binary Relabeling.} Find an optimal
  one-dimensional projection \(\alpha^{**}\), using a subset of
  \(\{(\mathbf{x_i},y_i^*)\}_{i=1}^n\) to separate the two class problem
  \(g_1^*\) and \(g_2^*\). To this second projection \textbf{only
  information from the two closest groups}, one from \(g_1^*\) and the
  other from \(g_2^*\) are used. The best separation of \(g_1^*\) and
  \(g_2^*\) is determined by
  \(\boldsymbol{\alpha}^{*T} \mathbf{\bar x}_{g1^*}< c\); then \(g_1^*\)
  is assigned to the left node else assign \(g_2^*\) to the right node,
  where \(\mathbf{\bar x}_{g1^*}\) is the mean of \(g_1^*\).
\end{itemize}

In Figure~\ref{fig-diag} this would correspond to dropping class \(g_3\)
from the current node. The projection is then computed using only
classes \(g_1\) and \(g_2\) (relabeled as \(g_1^*\) and \(g_2^*\) ). The
choice of the existing eight available rules is then applied to
determine the optimal split point, to create the terminal node for
\(g_1\). Subsequently, class \(g_3\) is reintroduced, and a new
projection is computed to best separate it from \(g_2\). The split is
made on this set of observations, creating two more terminal nodes,
completing the tree construction.

\subsection{Modification 2: Multiple splits per class based on entropy
reduction}\label{modification-2-multiple-splits-per-class-based-on-entropy-reduction}

\begin{figure}

\centering{

\includegraphics[width=1\linewidth,height=\textheight,keepaspectratio]{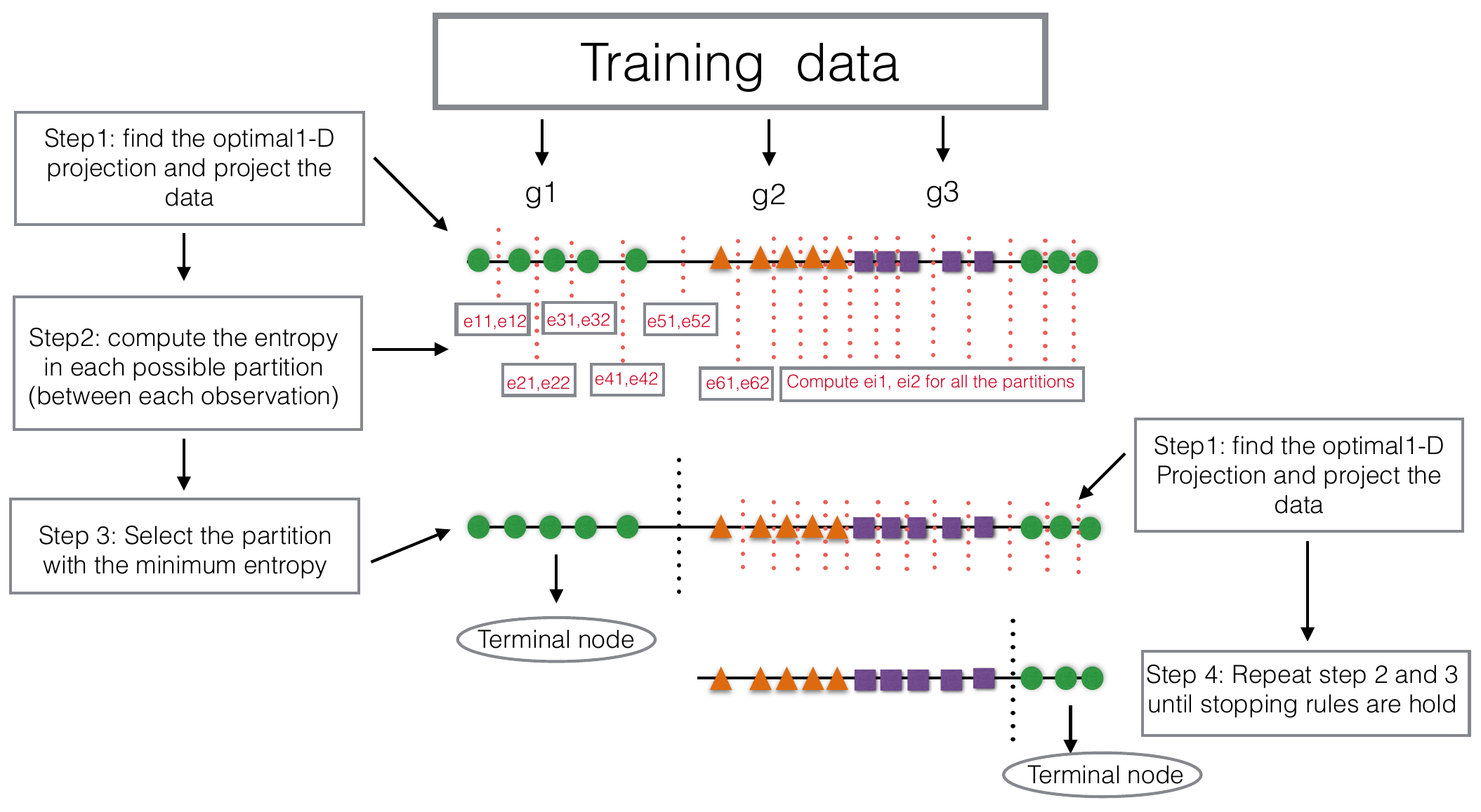}

}

\caption{\label{fig-diagMOD4}Illustration of the algorithm with
modification 2 applied to a three-class problem. Projections of the data
are computed at each node, and multiple splits per class are allowed. An
impurity-based criterion, such as entropy, is used to determine when and
how often to split observations within a node.}

\end{figure}

To allow for multiple splits, various changes are needed: the split is
made on the optimal projection, not after groups are combined, and new
stopping rules are defined to prevent over-fitting.
Figure~\ref{fig-diagMOD4} illustrates the change to the algorithm for
this modification.

In PPtree, the decision boundary split is made using the means for the
two groups, just like is done with linear discriminant analysis. This
modification changes this to follow the \texttt{rpart} algorithm,
minimizing entropy, with the difference being that it is done on the
optimal 1-D projected data. That is, the modification computes the
negative entropy criteria:

\begin{equation}
E(s) = -\sum_{j=1}^{G} p_{js} \log(p_{js})
\end{equation}

\noindent where \(p_{js}\) is the proportion of observations from class
\(j\) in subset \(s\), and \(G\) is the number of classes. Lower values
of \(E(s)\) indicate purity, reaching zero when all observations in the
subset belong to the same class. The impurity value for a split is
calculated by combining the values of the left and right subsets:

\begin{equation}
E(s_L, s_R) = \frac{n_L}{n_L + n_R} E(s_L) + \frac{n_R}{n_L + n_R} E(s_R)
\end{equation}

where \(n_L, n_R\) are the number of observations in the left and right
subsets, respectively. Across all possible splits, the lowest combined
impurity, \(E(s_L,s_R)\), is computed for all possible split points,
defined as the midpoints between consecutive ordered projected values,
denoted \(\mathscr{C} = \{c_1, ..., c_{n_L+n_R-1}\}\). The split point
that minimizes the combined entropy is selected:

\[{\min_{c\in \mathscr{C}}} E_c(s_L,s_R)\].

In addition, three stopping rules are introduced, which are similar to
\texttt{rpart}: the node is pure, there are fewer than a specified
number of observations in the node, or the reduction in impurity is less
than a specified tolerance value. These changes provide the modified
algorithm to be:

\textbf{Input:} Dataset \(d_n = \{(\mathbf{x}_i, y_i)\}_{i=1}^n\), with
\(\mathbf{x}_i \in \mathbb{R}^p\),
\(y_i \in \mathscr{G} = \{1, \ldots, G\}\)

\textbf{Output:} A projection pursuit classification tree with
entropy-based node splitting

\begin{itemize}
\item
  \textbf{Step 1: Class Separation.} Optimize a PP index (e.g., LDA or
  PDA) to find a projection vector \(\boldsymbol{\alpha}\) that best
  separates the classes in the current node.
\item
  \textbf{Step 2: Projection and Candidate Splits.} On the projected
  data, \(z_i = \boldsymbol{\alpha}^T \mathbf{x}_i\), define candidate
  splits as midpoints between consecutive \(z_i\) values.
\item
  \textbf{Step 3: Impurity Evaluation.} For each candidate split \(c\),
  divide the data into \(s_L = \{z_i < c\}\) and
  \(s_R = \{z_i \geq c\}\). Compute the class proportions \(p_{js}\) for
  each subset and evaluate the entropy:
\end{itemize}

\[E(s) = -\sum_{j=1}^G p_{js} \log(p_{js})\] Compute the total impurity
as:

\[E(s_L, s_R) = \frac{n_L}{n_L + n_R} E(s_L) + \frac{n_R}{n_L + n_R} E(s_R)\]

\begin{itemize}
\item
  \textbf{Step 4: Best Split Selection.} Select the split point \(c^*\)
  that minimizes \(E(s_L, s_R)\). Split the node using the rule
  \(z_i < c^*\).
\item
  \textbf{Step 5: Recursive Partitioning.} Repeat Steps 1--4 recursively
  on each child node until some of the following stopping criteria are
  met:

  \begin{itemize}
  \tightlist
  \item
    All observations in the node belong to the same class;
  \item
    The node size is smaller than a minimum threshold \(n_s\);
  \item
    The entropy reduction is smaller than a threshold \(ent_s\).
  \end{itemize}
\end{itemize}

\section{Comparative performance}\label{sec-compare}

Three approaches are used to examine the effectiveness of the
modifications. The first is a Shiny app which generates structures in
2-D and shows the model fits for three algorithms: \texttt{rpart},
PPtree, and the PPtree extensions. It was developed and used to decide
on effective changes to the algorithm. The second is a simulation study
on 94 benchmark data sets to compare test errors. Lastly, to understand
where, why, and how the extensions work, we examine the data visually in
high dimensions along with the fitted models.

\subsection{Exploring the algorithm changes
interactively}\label{sec-algocomp}

The Shiny app includes methods for simulating 2D classification
problems, and displays the decision boundaries generated by three
different algorithms: CART, PPtree, and the proposed extensions as
implemented in the \texttt{rpart}, \texttt{PPtreeViz}, and
\texttt{PPtreeExt} packages, respectively.

\begin{figure}[t]

\centering{

\includegraphics[width=1\linewidth,height=\textheight,keepaspectratio]{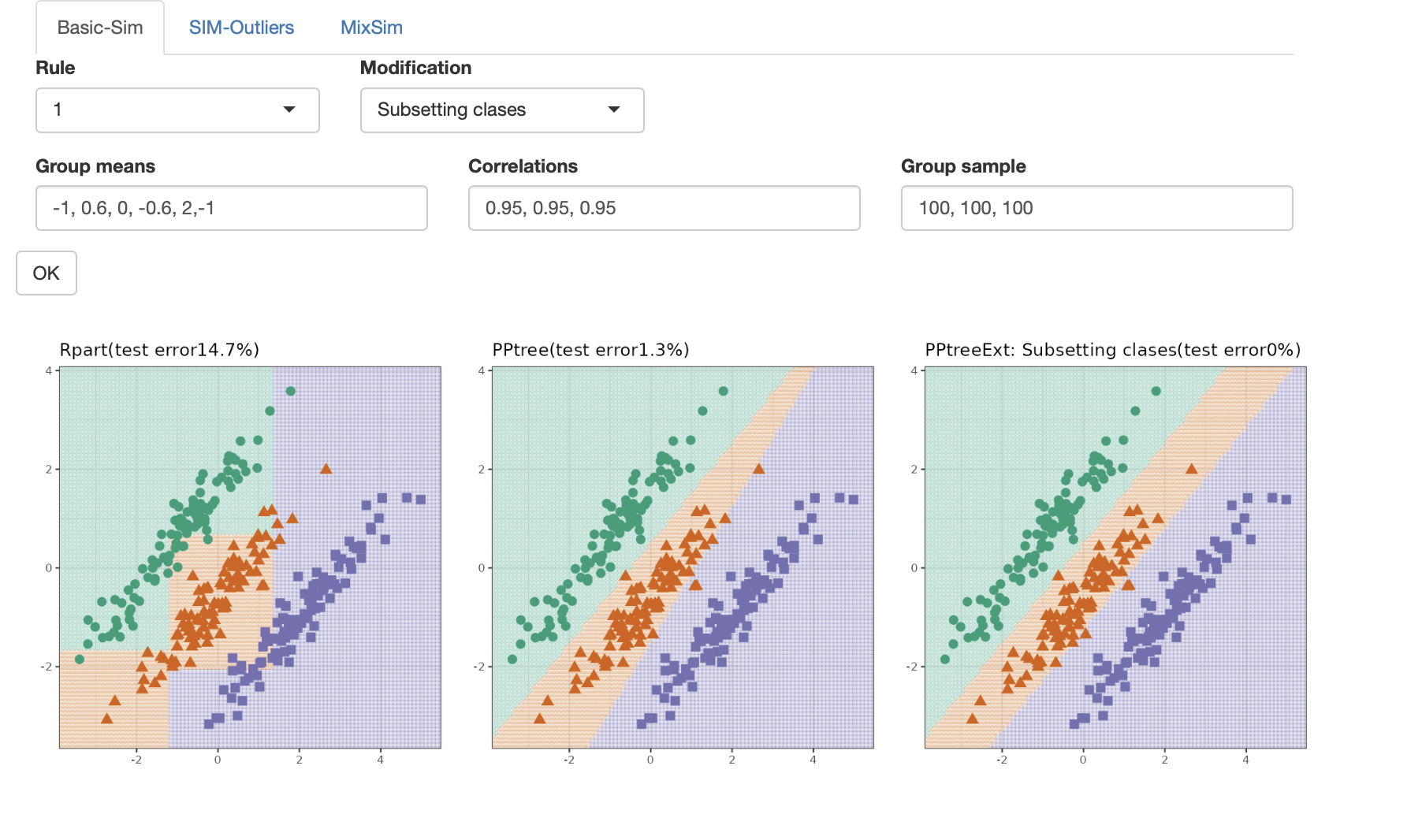}

}

\caption{\label{fig-tab1}Comparing the fits for three-class data with a
strong correlation between variables. Modification 1 (different
subsetting rules) of \texttt{PPtree} is shown. The modification produces
lower error and decision boundaries that are more symmetrically placed
between the class clusters.}

\end{figure}

\begin{figure}[t]

\centering{

\includegraphics[width=1\linewidth,height=\textheight,keepaspectratio]{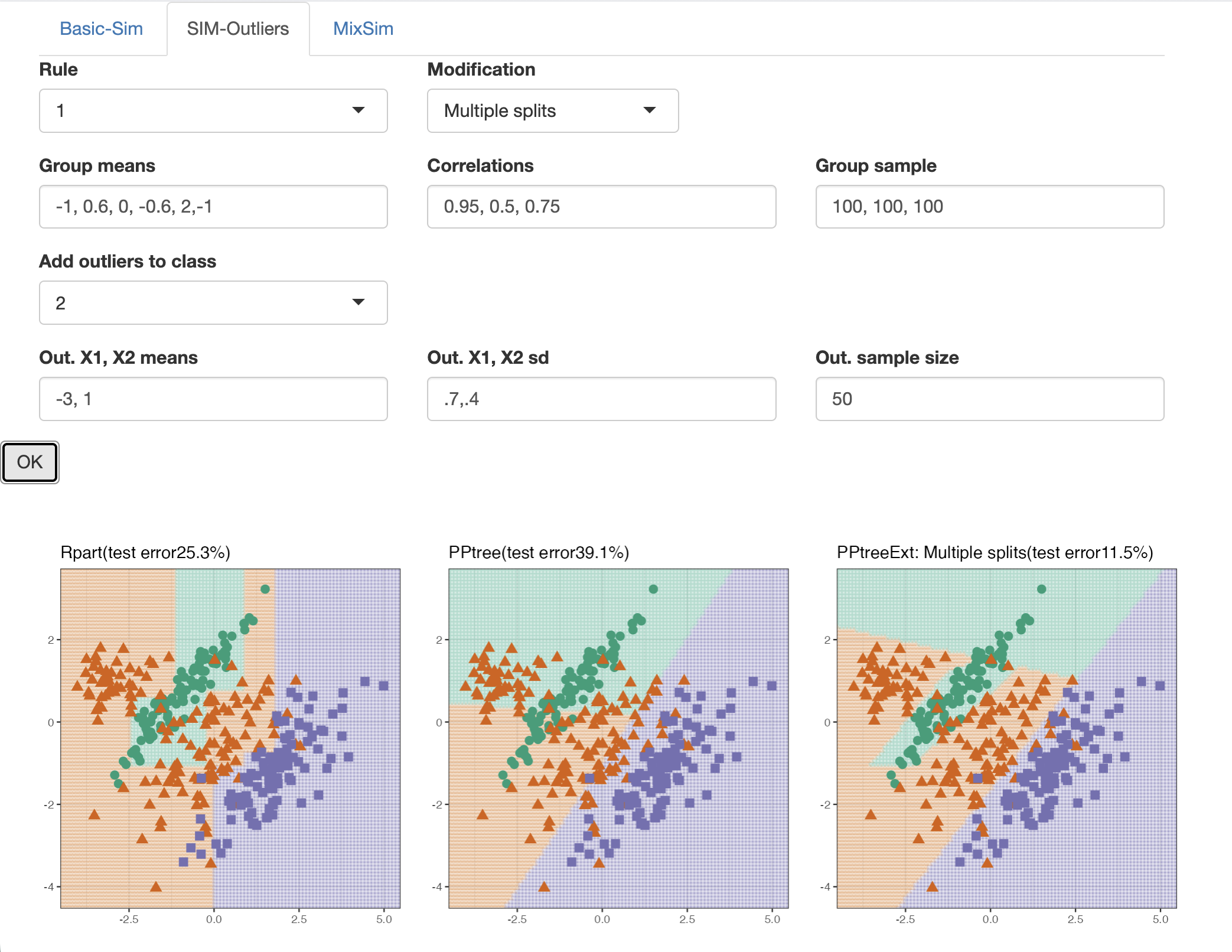}

}

\caption{\label{fig-tab2}Comparing the fits for data where one class has
two distinct clusters. Modification 2 (multiple splits) of
\texttt{PPtree} is used. The modification produces lower error and
better decision boundaries.}

\end{figure}

\begin{figure}[t]

\centering{

\includegraphics[width=1\linewidth,height=\textheight,keepaspectratio]{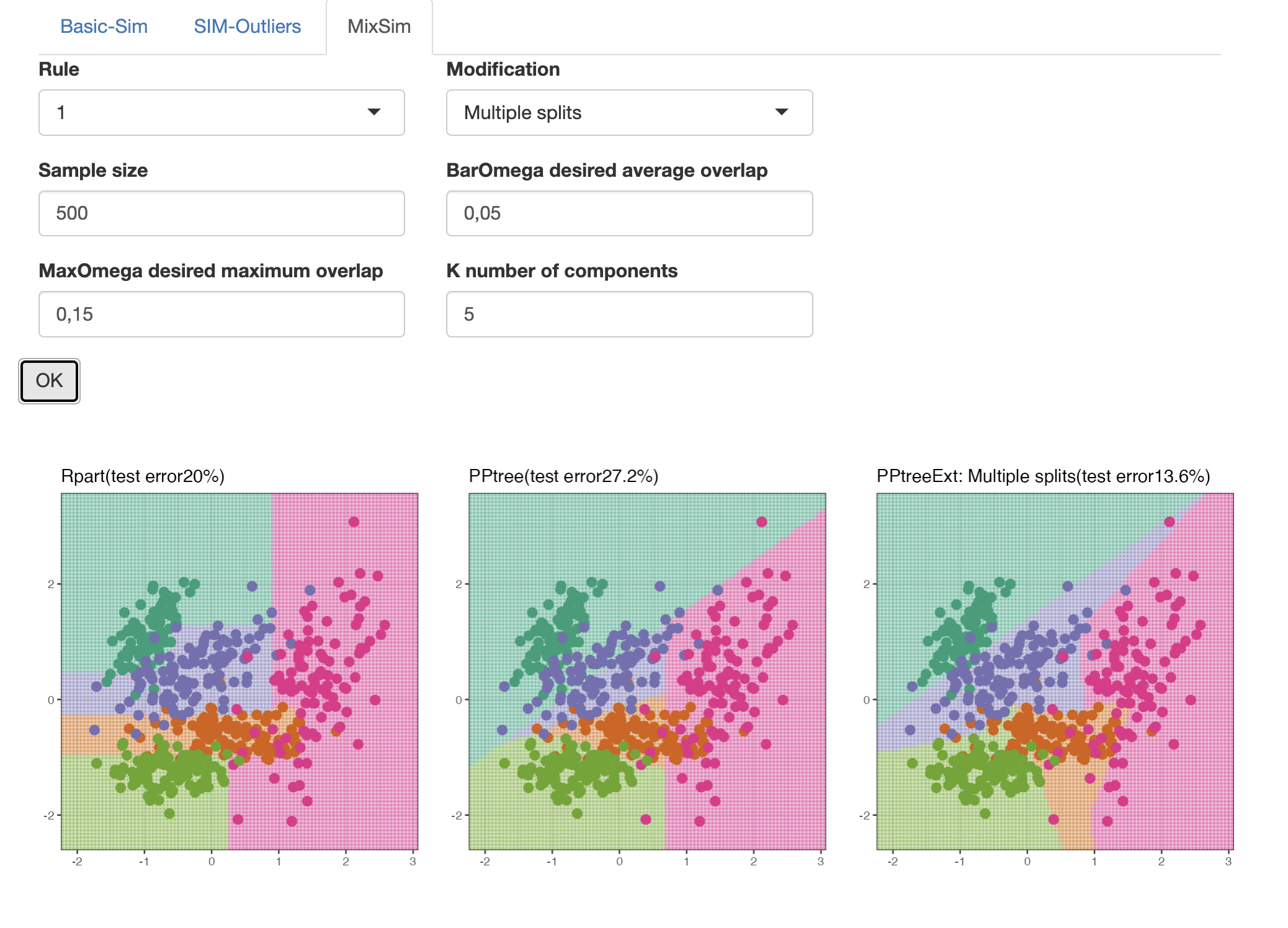}

}

\caption{\label{fig-tab3}Comparing the fits for five-cluster Gaussian
mixtures. Modification 2 of \texttt{PPtree} is used. This produces lower
error and reasonable decision boundaries.}

\end{figure}

There are three data generation methods, ``Basic-Sim'', ``Sim-Outlier'',
and ``MixSim''. Basic-Sim simulates classes with separations between
clusters on linear combinations of variables, by user-specified
correlations between variables for each cluster. Sim-Outlier uses the
same approach with the addition of a cluster of outliers corresponding
to one of the classes. These two methods allow only for three-cluster
data. The user can control the number of observations in each class, the
class means, and the Sim-Outlier mean and shape of the outlier groups.
MixSim uses a Gaussian mixture model using the \texttt{MixSim} package
(Melnykov et al. 2012) to simulate class clusters, with any number of
classes. While all classes have equal sample size, the user can choose
this and also specify values for the \texttt{MixSim} algorithm.
Figure~\ref{fig-tab1}, Figure~\ref{fig-tab2}, Figure~\ref{fig-tab3} show
the interface for the three different data-generating methods. Clicking
``Ok'' will simulate new samples.

The decision boundaries for \texttt{rpart}, \texttt{PPtree}, and the
modified \texttt{PPtree} (implemented in \texttt{PPtreeExt}) are
displayed side by side for visual comparison. The user can choose which
modifications of PPtree to show. For the other algorithms, only the
default parameter choices are used.

In Figure~\ref{fig-tab1}, three-class data with a strong correlation
between variables is simulated. Modification 1 (different subsetting
rules) is applied, and rule 1 (Table~\ref{tbl-rules}) is used to compute
the split point. The modified \texttt{PPtree} achieves an error rate
comparable to the original \texttt{PPtree}, with both substantially
outperforming \texttt{rpart}. The change in boundaries is subtle, but
more satisfying. Between classes S (violet squares) and T (orange
triangles), the boundary better matches the separation - it is not
pulled away from this split by using LDA on the super-group (T and C).

Figure~\ref{fig-tab2} shows data where one class has two separated
clusters, as simulated by Sim-Outlier. The PPtree extension used is
modification 2 (multiple splits). It yields a substantial improvement
over the other algorithms, both in lower error and more sensible
boundaries. The use of multiple splits enables the small group of
outliers from class T (orange triangles) to be better classified.

The data in Figure~\ref{fig-tab3} has 5 clusters of different shapes.
Modification 2 (multiple splits) is examined. Again, we see that the
modified algorithm outperforms PPtree and \texttt{rpart}, in both lower
error and better boundaries.

The interactive application provides insight into how modifications to
the algorithm affect model fit. Comparing the fitted models follows the
principle of visualizing the model in the data space (Wickham et al.
2015). This app has been a valuable tool for debugging and evaluating
the algorithmic behavior. By adjusting the simulated data and model
parameters, it becomes evident that small changes in the data can result
in substantial changes in the fitted model, particularly when
projections are involved. By understanding the effect in 2-D, we aim to
better understand how the algorithm might also behave in
high-dimensional settings.

\subsection{Numerical performance}\label{sec-result}

\begin{figure}[t]

\centering{

\pandocbounded{\includegraphics[keepaspectratio]{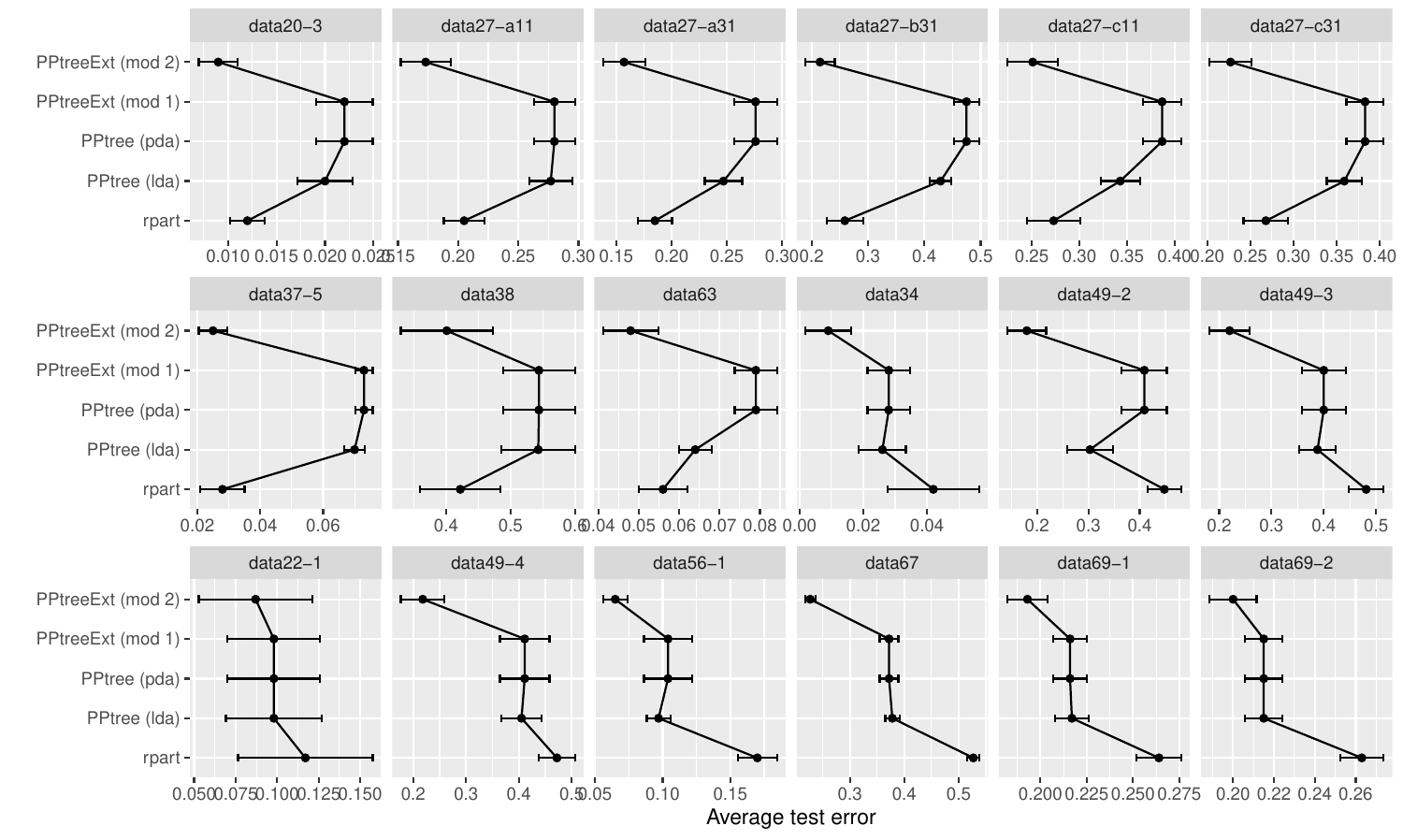}}

}

\caption{\label{fig-performance}Average test error with 95\% error bars
from 200 training/test splits of benchmark data for four models:
\texttt{rpart}, PPtree, \texttt{PPtreeExt} modifications 1 and 2, where
the latter performs better than the others. The order of the plots is
based on the pattern of errors: at top left \texttt{PPtreeExt} (mod 2)
is best but only slightly better than \texttt{rpart}, and at bottom
right each of the PPtree versions iteratively performs better than
\texttt{rpart}.}

\end{figure}

We compare the predictive performance for the extended PPtree algorithm
with several classification methods, CART, PPtree (LDA, PDA), random
forest, PPforest, and SVM, using 94 benchmark data sets. (LDA indicates
PP conducted with the linear discriminant index, and PDA indicates PP
conducted using the penalized linear discriminant index.)
Figure~\ref{fig-performance} summarizes the performance on data sets
where the modified algorithm outperformed other tree methods, and
\textbf{?@tbl-data-tab2} summarizes the main characteristics of each
data set, including the number of observations, predictors, number of
classes, class imbalance, and absolute average correlation among
predictors. Imbalance quantifies the disparity in class sizes (0
indicates perfectly balanced classes, the biggest proportion size
difference between classes), while higher correlation suggests a greater
potential for class separation through linear combinations of variables.
(The Appendix contains the links to the original data sources, full
performance, and data set summaries.)

In the simulation, two-thirds of the observations in each data set are
randomly selected for training (using stratified sampling), and the
remaining one-third is used for testing and computing the prediction
error. This process is repeated 200 times to assess the variance in
performance.

Figure~\ref{fig-performance} summarises the performance for the 18 data
sets, where \texttt{PPtreeExt} with modification 2 outperforms the other
tree algorithms. In three of these data sets, it also has the lowest
error than the ensemble-based classifiers, also. The plots are ordered
according to the pattern of test error. In data sets \texttt{data20-3},
\texttt{data27-a11}, \texttt{data27-a31}, \texttt{data27-b31},
\texttt{data27-c11}, \texttt{data27-c31}, \texttt{data37-5},
\texttt{data38}, and \texttt{data63}, only \texttt{PPtreeExt} (mod 2)
outperforms \texttt{rpart}. For \texttt{data34} and \texttt{data49-2}
results are mixed. For the other data sets (\texttt{data49-3},
\texttt{data22-1}, \texttt{data49-4}, \texttt{data56-1},
\texttt{data67}, \texttt{data69-1}, \texttt{data69-2}), each of the
methods does better than \texttt{rpart}. The variability across training
and test splits is high for some data sets (\texttt{data38},
\texttt{data22-1}). Data sets \texttt{data20-3}, \texttt{data49-2}, and
\texttt{data69-1}, representing a range of numerical performance, are
chosen to investigate why \texttt{PPtreeExt} performs better, using
high-dimensional visualisation in the next section.

\begin{longtable}[]{@{}lrrrrr@{}}
\caption{Overview of benchmark datasets where the proposed tree-based
extensions outperform the selected tree methods. The table summarizes
the number of observations, predictors, number of classes, class
imbalance, and absolute average correlation among predictors. Imbalance
quantifies the disparity in class sizes (0 indicates perfectly balanced
classes), while higher correlation suggests a greater potential for
class separation through linear combinations of
variables.}\label{tbl-datatbl}\tabularnewline
\toprule\noalign{}
dataset & Cases & Predictors & Groups & Imbalance & Correlation \\
\midrule\noalign{}
\endfirsthead
\toprule\noalign{}
dataset & Cases & Predictors & Groups & Imbalance & Correlation \\
\midrule\noalign{}
\endhead
\bottomrule\noalign{}
\endlastfoot
data20-3 & 8143 & 5 & 2 & 0.58 & 0.45 \\
data27-a11 & 1747 & 18 & 5 & 0.38 & 0.24 \\
data27-a31 & 1834 & 18 & 5 & 0.28 & 0.27 \\
data27-b31 & 1424 & 18 & 5 & 0.21 & 0.24 \\
data27-c11 & 1111 & 18 & 5 & 0.14 & 0.27 \\
data27-c31 & 1448 & 18 & 5 & 0.17 & 0.27 \\
data37-5 & 17560 & 53 & 5 & 0.21 & 0.16 \\
data38 & 182 & 12 & 2 & 0.43 & 0.21 \\
data63 & 6598 & 166 & 2 & 0.69 & 0.27 \\
data34 & 1372 & 4 & 2 & 0.11 & 0.43 \\
data49-2 & 606 & 100 & 2 & 0.03 & 1.00 \\
data49-3 & 606 & 100 & 2 & 0.01 & 0.99 \\
data22-1 & 258 & 5 & 4 & 0.25 & 0.12 \\
data49-4 & 606 & 100 & 2 & 0.01 & 0.99 \\
data56-1 & 7494 & 16 & 10 & 0.01 & 0.27 \\
data67 & 20000 & 16 & 26 & 0.00 & 0.18 \\
data69-1 & 5000 & 21 & 3 & 0.01 & 0.30 \\
data69-2 & 5000 & 40 & 3 & 0.01 & 0.09 \\
\end{longtable}

\subsection{Visual explanations of performance}\label{sec-perfexp}

Exploring high dimensions is tricky, so it is helpful to set some
objectives before leaping into the vast space. We would expect the
modified PPtree algorithm to

\begin{itemize}
\tightlist
\item
  outperform \texttt{rpart} when the best separation between groups is
  in oblique directions
\item
  and outperform PPtree when class variances are different, or when
  there are multiple clusters for any class.
\end{itemize}

This can be investigated visually using a tour. There are several
strategies to use for approaching this:

\begin{enumerate}
\def\labelenumi{\arabic{enumi}.}
\tightlist
\item
  On the data

  \begin{enumerate}
  \def\labelenumii{\alph{enumii}.}
  \tightlist
  \item
    Work on two groups at a time, if there are more than 2
  \item
    Use a guided tour on the data to find the combinations of the
    variables where the groups are separated
  \end{enumerate}
\item
  Predict a grid of values that covers the data domain, for the methods
  interested in comparing, \texttt{PPtreeExt}, \texttt{rpart}, and
  PPtree, as can be done using the \texttt{classifly} package (Wickham
  2022), and then

  \begin{enumerate}
  \def\labelenumii{\alph{enumii}.}
  \tightlist
  \item
    subset to the points near the border generated by the classifier.
  \item
    record a sequence of projection bases that shows the boundary
    created by the method in focus.
  \item
    use the same projection sequence to view the boundaries for the
    three different methods, and possibly use a slice tour to focus on
    the boundary near the center of the data domain.
  \end{enumerate}
\end{enumerate}

\begin{figure}

\begin{minipage}{0.50\linewidth}

\includegraphics[width=3.02083in,height=\textheight,keepaspectratio]{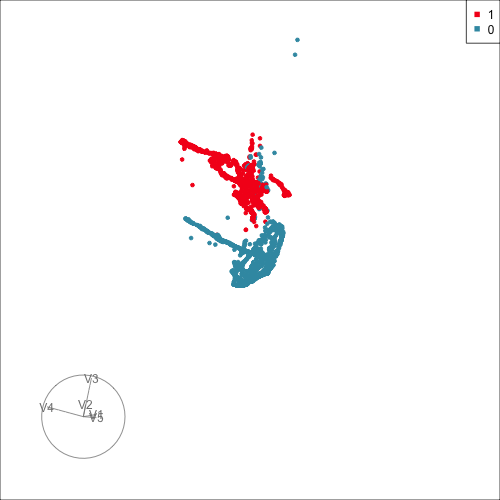}

\subcaption{\label{}projection 1}
\end{minipage}%
\begin{minipage}{0.50\linewidth}

\includegraphics[width=3.02083in,height=\textheight,keepaspectratio]{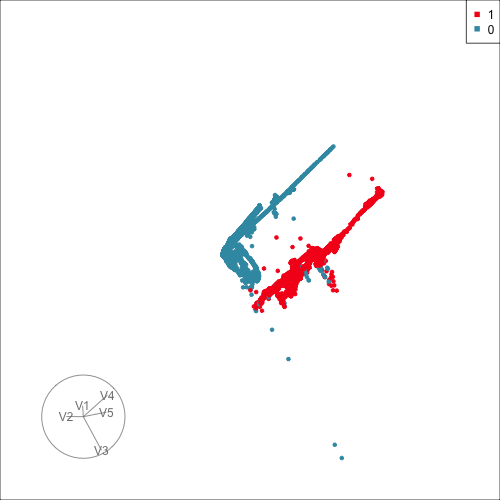}

\subcaption{\label{}projection 2}
\end{minipage}%

\caption{\label{fig-data20-3}Two projections from a grand tour of
data20-3, from which we can see why the \texttt{PPtreeExt} method is the
most effective. Class 0 is split into two clusters on either side of
group 1, which means that the original algorithm is not optimal.
Secondly, the separation between the two groups is primarily \texttt{V3}
with small contributions from other variables, which gives
\texttt{PPtreeExt} a slight edge over \texttt{rpart}.}

\end{figure}

Figure~\ref{fig-data20-3} shows two projections of the \texttt{data20-3}
illustrating why \texttt{PPtreeExt} performs better. The projections are
2-D from the full 5-D space. Color indicates the two classes (0 is blue
and 1 is red). The circles represent the projection coefficients. Using
a clock dial analogy, in the first projection (a), \texttt{V3} is
pointing towards 1 o'clock, and \texttt{V4} is pointing towards 10.
\texttt{V2} is also pointing to 1, but is very small. The difference in
the two groups is in the direction of 1, so \texttt{V3} is likely mostly
contributing to this, with a small contribution from \texttt{V2}. In the
second projection (b) \texttt{V3} points to 5, \texttt{V4} points to 2,
\texttt{V5} points to 3, and \texttt{V2} points to 9. The separation is
in the direction of 5 o'clock, suggesting again that \texttt{V3} is
primarily important. Group 0 is split on either side of group 1, which
means that the PPtree algorithm would not be able to capture both
clusters, but modification 2 would. The separation between the two
classes is mostly on \texttt{V3} but small contributions from other
variables help to improve the separation. This means that \texttt{rpart}
will not work quite as well because it will focus on splitting
\texttt{V3}, ignoring the other variables.

\begin{figure}

\begin{minipage}{0.33\linewidth}

\includegraphics[width=2.08333in,height=\textheight,keepaspectratio]{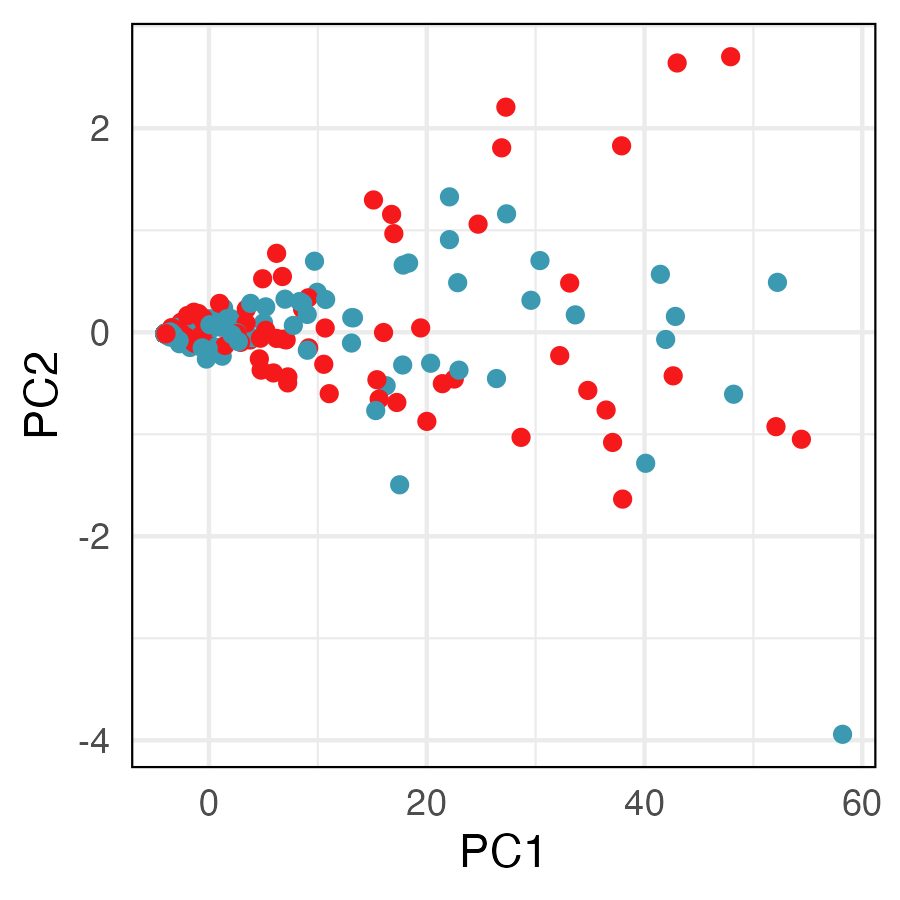}

\subcaption{\label{}PC1 vs PC2}
\end{minipage}%
\begin{minipage}{0.33\linewidth}

\includegraphics[width=2.08333in,height=\textheight,keepaspectratio]{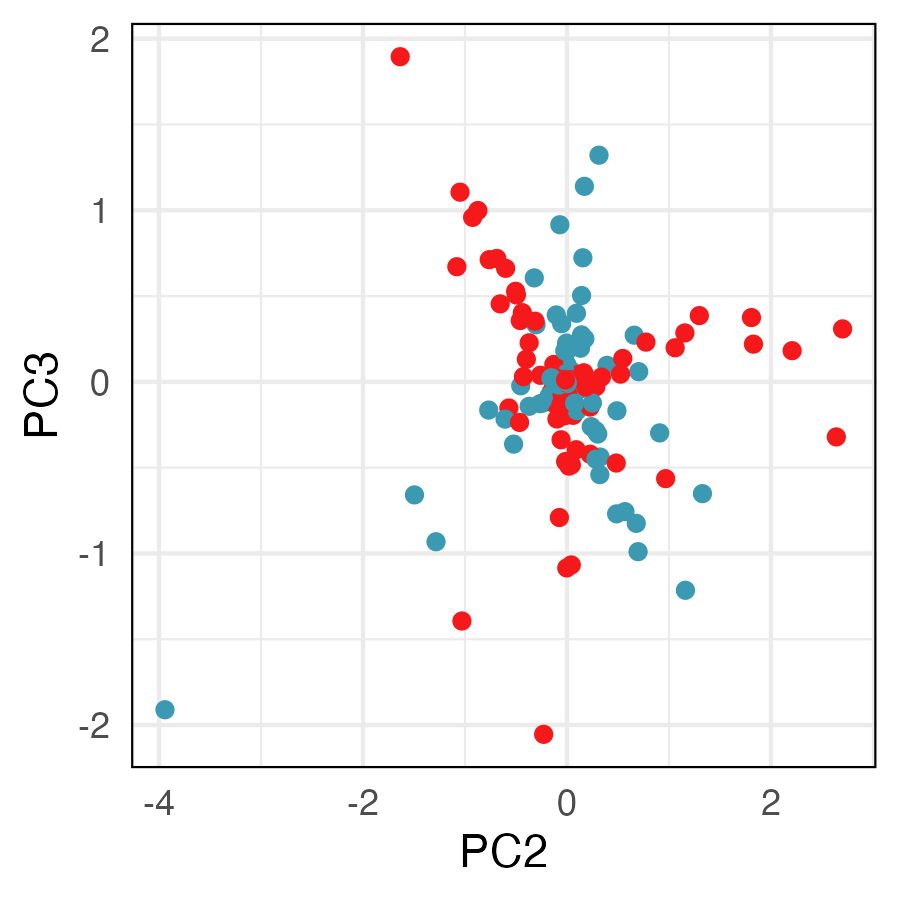}

\subcaption{\label{}PC2 vs PC3}
\end{minipage}%
\begin{minipage}{0.33\linewidth}

\includegraphics[width=2.08333in,height=\textheight,keepaspectratio]{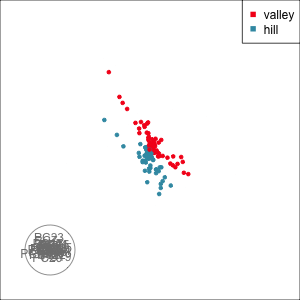}

\subcaption{\label{}optimal projection}
\end{minipage}%

\caption{\label{fig-data49-2}Examining why \texttt{data49-2} is best
classified using \texttt{PPtreeExt}. Because there are 100 predictors,
for visualisation it is useful to reduce the dimension using principal
component analysis, and then examine the separation between groups. PC1
explains 99.8\% of the overall variability among predictors, but it is
not useful for separating the classes (a). PC2 and PC3 (b) have an
interesting difference between the two classes, which shows why multiple
linear splits might be useful. A guided projection pursuit tour on the
first 26 principal components (c) shows that these classes could be
separated with a linear combination. This is why the original PPtree
classifier does better than the tree algorithm and also why the
extension improves the performance.}

\end{figure}

From Figure~\ref{fig-performance}, for \texttt{data49-2}, we can see
that the original PPtree (with LDA) algorithm performs much better than
a tree, and the \texttt{PPtreeExt} performs much better again. This data
has 100 predictors and 2 classes (Table~\ref{tbl-datatbl}), which makes
visualising more challenging than \texttt{data20-3}. A useful approach
is to pre-process using principal component analysis (PCA) on the
predictors, to reduce the number of variables to visualise, while
maintaining class structure. In Figure~\ref{fig-data49-2} plot (a) shows
PC 1 vs PC2, with points coloured by the two classes. PC 1 explains
99.8\% of the variability in the predictors, but it explains nothing of
the class structure. This is not uncommon for PCA, where an interesting
structure can be found in the smaller components. Plot (b) shows PC 2 vs
PC 3, and here we see interesting class differences. Each class appears
to have three linear pieces pointing in different directions. This
supports why \texttt{PPtreeExt} might work well: in the combination of
variables, the classes are nearly separable with multiple linear cuts.
It's not feasible to continue working through principal components to
tease apart the cluster structure relating to classes, so a good next
step is to do PP with the guided tour. The optimal projection is shown
in plot (c). It is computed on 26 PCs, and many of these contribute to
this projection as seen by the mess of lines inside the circle
representing the axes. The two classes are mostly linearly separable in
this projection, but there is no clear gap between the clusters. The
linear combination with so many variables is what causes some
difficulties for a simple tree algorithm. The interesting shapes of the
clusters for each class, which appear to be multiple linear pieces, mean
that the \texttt{PPtreeExt} can find better separations than PPtree.

From Figure~\ref{fig-performance} we see that all of the PPtree versions
perform better than an \texttt{rpart} fit on \texttt{data69-1}. This
data has 21 predictors and 3 classes (Table~\ref{tbl-datatbl}). This is
a sufficiently small-scale problem to examine using a tour, and we
initially use a guided tour with the LDA index. This produced the
projection shown in plot (a), where we see the primary difference
between the three classes. These are each similar size, but oriented in
such a way that in this 2-D projection, they form a triangle. Plots (b)
and (c) show the boundaries using a slice tour with the same projection
as used in (a) for \texttt{PPtreeExt} and \texttt{rpart}, respectively.
This uses approach 2 above: created by predicting the class of points
randomly generated from a uniform distribution in a 21-D cube of the
same range as the data. The slice cuts through to the center of the box.
We can see that \texttt{PPtreeExt} essentially fits the data shape
neatly, but the \texttt{rpart} boundary is very messy. This data needs
oblique cuts, which \texttt{PPtreeExt} provides but \texttt{rpart} does
now. The modified algorithm handles the variance differences better than
the original PPtree.

\begin{figure}

\begin{minipage}{0.33\linewidth}

\includegraphics[width=1.875in,height=\textheight,keepaspectratio]{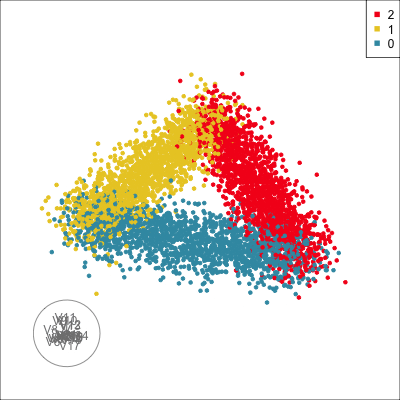}

\subcaption{\label{}data}
\end{minipage}%
\begin{minipage}{0.33\linewidth}

\includegraphics[width=1.875in,height=\textheight,keepaspectratio]{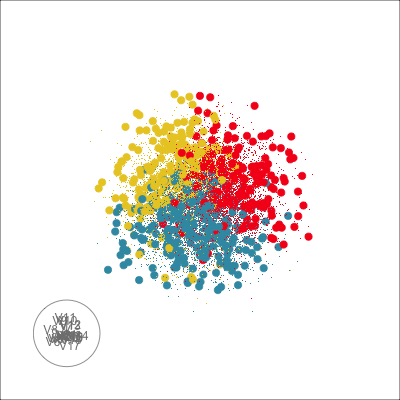}

\subcaption{\label{}PPtreeExt}
\end{minipage}%
\begin{minipage}{0.33\linewidth}

\includegraphics[width=1.875in,height=\textheight,keepaspectratio]{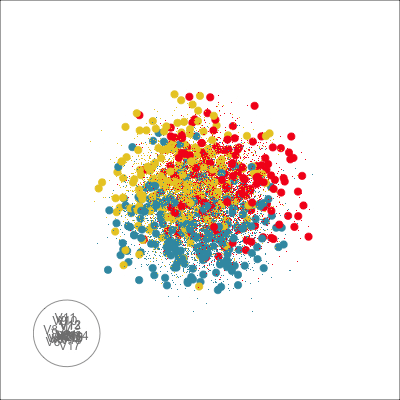}

\subcaption{\label{}rpart}
\end{minipage}%

\caption{\label{fig-data69-1}Examining why \texttt{data69-1} is best
classified using \texttt{PPtreeExt}: (a) best projection from a guided
tour shows that the data has three similar linear pieces oriented to
form a triangle, (b) boundary produced by the \texttt{PPtreeExt} (mod 2)
fitted model, and (c) boundary produced by \texttt{rpart} fitted model.
The same projection is used in all plots to allow comparison. The
boundary is examined by predicting the class of a uniform random sample
of points in the data space, and viewed using a slice tour. The boundary
for \texttt{PPtreeExt} (mod 2) mirrors what we learn from the data, that
it is carving the data space into three equiangular chunks, but
\texttt{rpart} is not capturing these linear combinations.}

\end{figure}%

\section{Discussion}\label{sec-disc}

This article has presented two modifications to the \texttt{PPtree}
algorithm for classification problems implemented in the R package
\texttt{PPtreeExt}. These changes result in a more flexible
classification method that leverages linear combinations of variables.
The performance study showed that the predictive performance of
\texttt{PPtreeExt} with modification 2 is better than CART and PPtree in
18 out of 94 data sets. A web app was developed and shared to help
understand and debug modifications to the algorithm.

Interpretability is increasingly important because predictions from
algorithms may need to be defended in practice. We have shown how to
examine the data and the fitted model using high-dimensional visual
methods to understand why the PPtree extension performs better than
other tree methods on some data sets. This approach is generally useful
for assessing and diagnosing model fits.

Ultimately, this may provide better building blocks for a projection
pursuit-based classification random forest algorithm (da Silva et al.
2021) as available in the \texttt{PPforest} R package (da Silva et al.
2025). The modified tree algorithm can be incorporated into a projection
pursuit forest classifier (da Silva et al. 2025). Given its implicit
variable selection mechanism, the ensemble is expected to handle
nuisance variables more effectively than a single tree. Using the PDA
index (penalised discriminant analysis) also helps when there are many
variables. Additionally, the use of bagging should yield improved and
more robust performance in problems with nonlinear decision boundaries.
This can be accomplished using the current PPforest package.

\section{Supplementary Materials}\label{supplementary-materials}

This article was written in Quarto (Allaire et al. 2024), using the R
package \texttt{ggplot2} (Wickham 2016) for visualizations. The files to
reproduce the article and results are available at
\url{https://github.com/natydasilva/PPtreeExt_paper_JCGS}. Animated gifs
corresponding to Figure~\ref{fig-data20-3}, Figure~\ref{fig-data49-2}
and Figure~\ref{fig-data69-1} are in the \texttt{gifs} folder. The
\texttt{code} directory contains the scripts for the simulations and to
explore the model fits in high dimensions.

The \texttt{PPtreeExt} package provides the modified PPtree algorithm.
The Shiny app is available at
\url{https://natyds.shinyapps.io/explorapp/} and the app code is
provided in the \texttt{PPtreeExt} package, so that it can be used
locally for new algorithm developments.

\newpage

\phantomsection\label{refs}

\newpage
\subsection{Appendix}\label{appendix}

\begin{figure}

\centering{

\pandocbounded{\includegraphics[keepaspectratio]{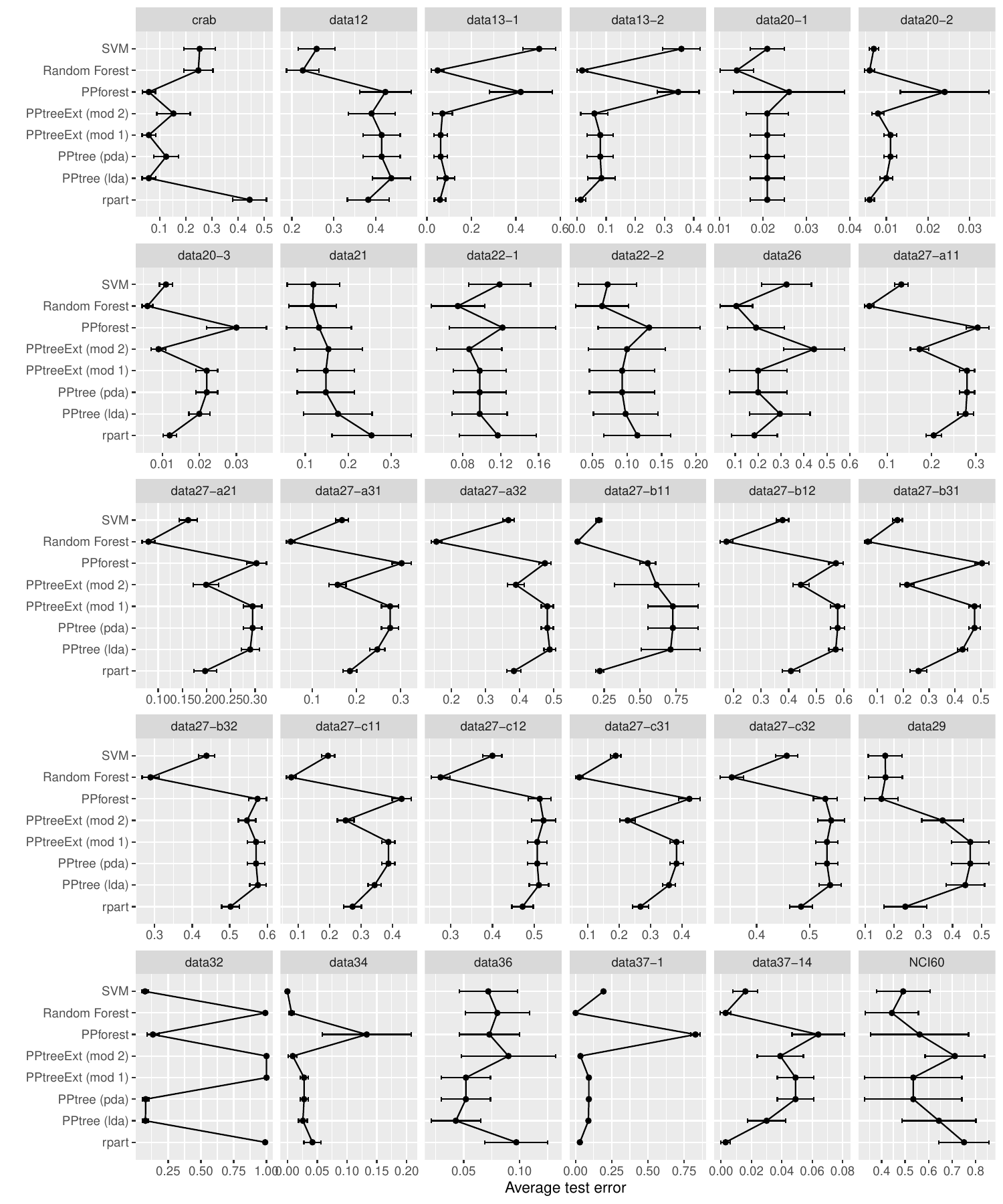}}

}

\caption{\label{fig-Figure13}Average test error from 200 training/test
splits of benchmark data for: SVM, Random Forest, PPtree (pda), PPtree
(LDA), \texttt{PPtreeExt} modifications 1 and 2, \texttt{PPforest} and
\texttt{rpart}.}

\end{figure}%

\begin{figure}

\centering{

\pandocbounded{\includegraphics[keepaspectratio]{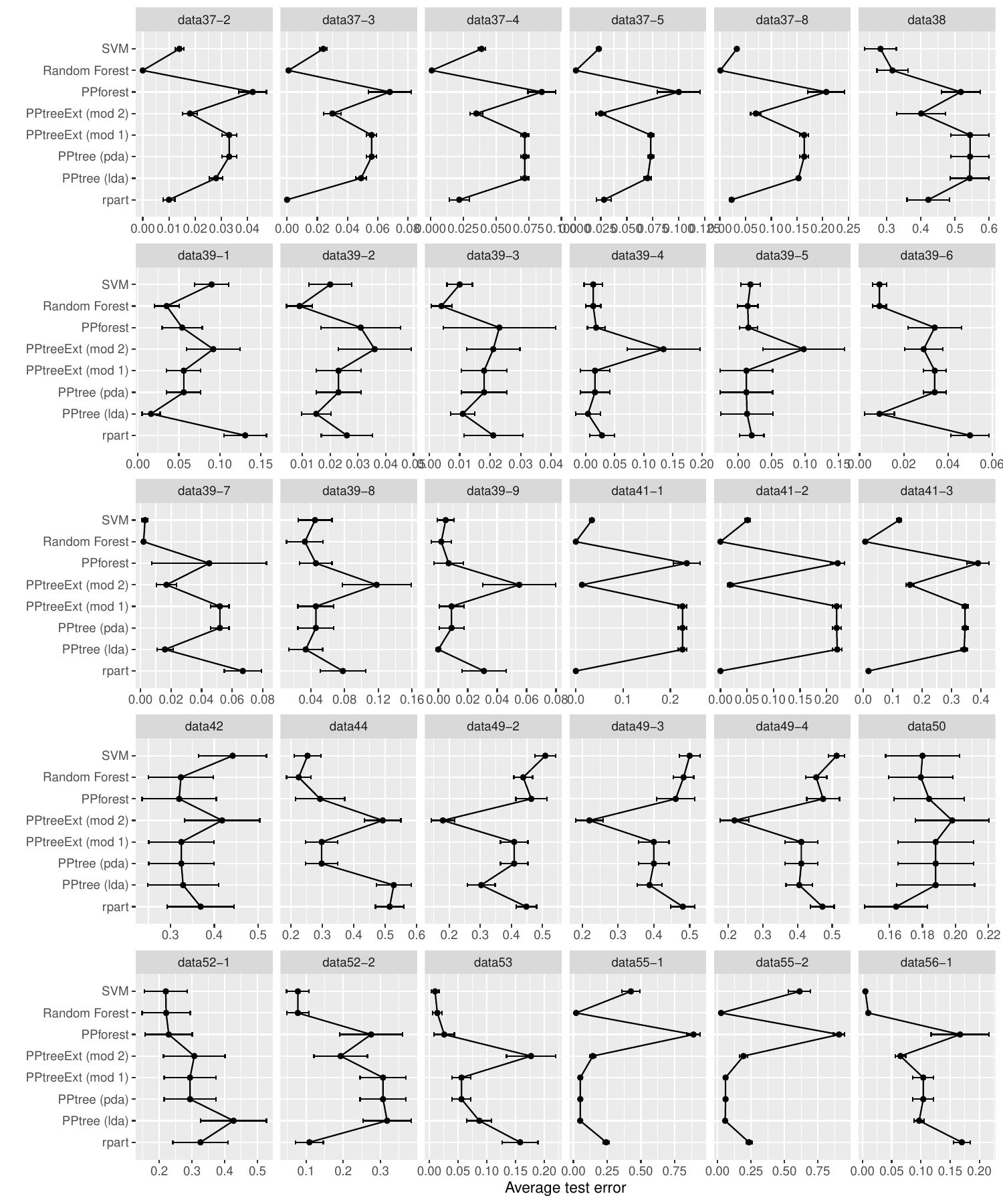}}

}

\caption{\label{fig-Figure14}Average test error from 200 training/test
splits of benchmark data for: SVM, Random Forest, PPtree (pda), PPtree
(LDA), \texttt{PPtreeExt} modifications 1 and 2, \texttt{PPforest} and
\texttt{rpart}.}

\end{figure}%

\begin{figure}

\centering{

\pandocbounded{\includegraphics[keepaspectratio]{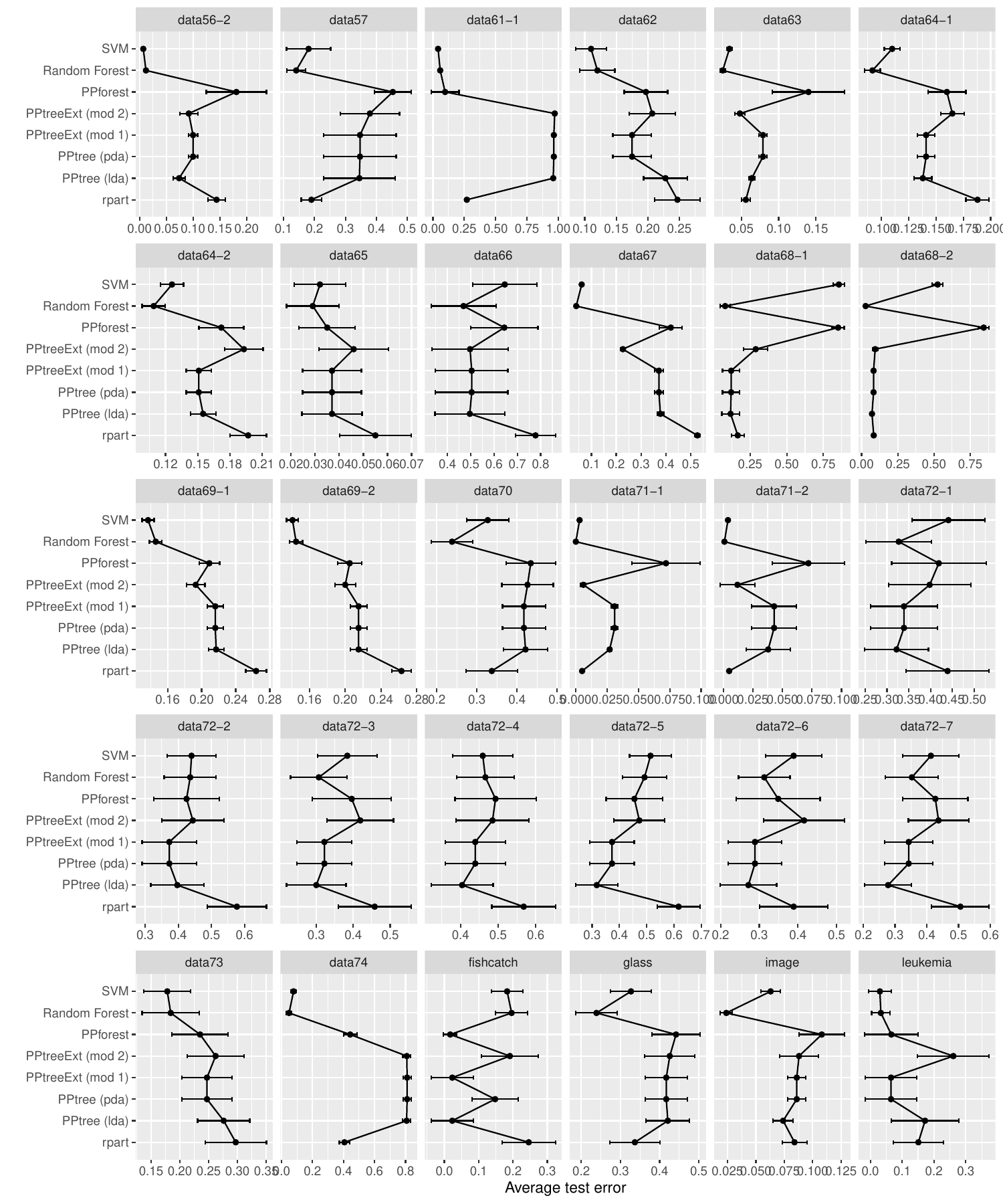}}

}

\caption{\label{fig-Figure15}Average test error from 200 training/test
splits of benchmark data for: SVM, Random Forest, PPtree (pda), PPtree
(LDA), \texttt{PPtreeExt} modifications 1 and 2, \texttt{PPforest} and
\texttt{rpart}.}

\end{figure}%

\begin{figure}

\centering{

\pandocbounded{\includegraphics[keepaspectratio]{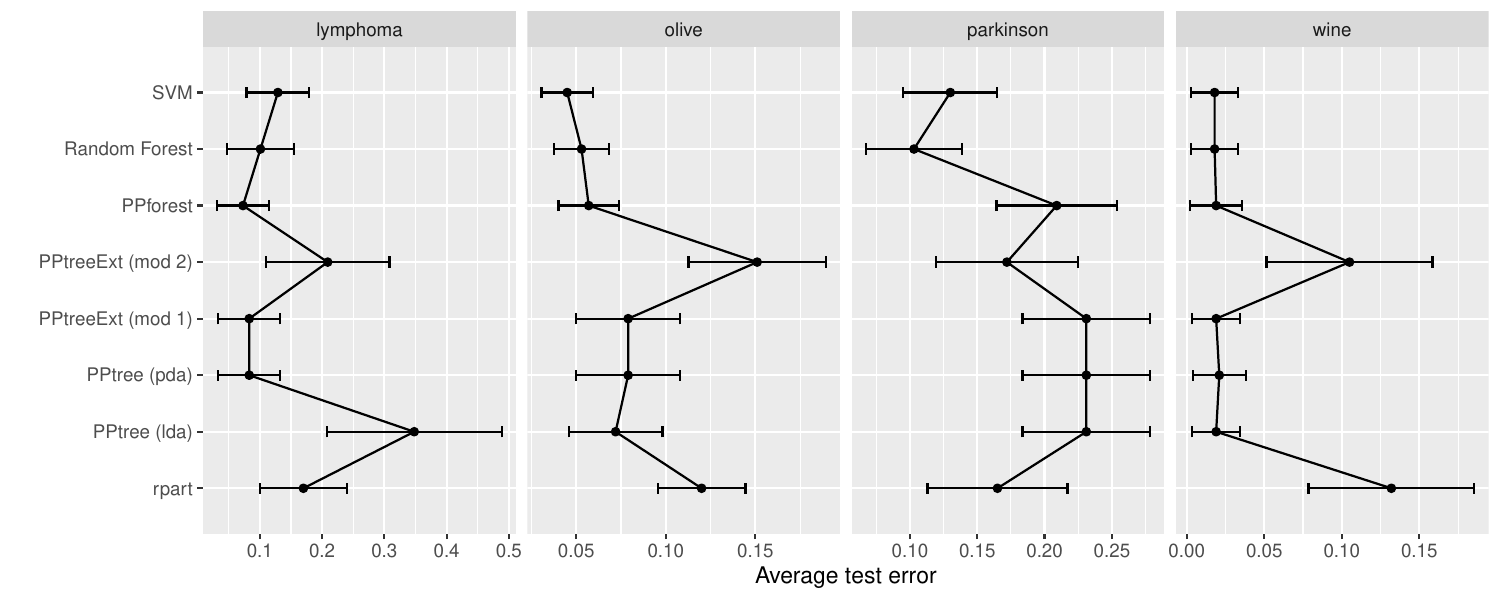}}

}

\caption{\label{fig-Figure16}Average test error from 200 training/test
splits of benchmark data for: SVM, Random Forest, PPtree (pda), PPtree
(LDA), \texttt{PPtreeExt} modifications 1 and 2, \texttt{PPforest} and
\texttt{rpart}.}

\end{figure}%

\begingroup\fontsize{10}{12}\selectfont

\begin{longtable}[t]{ll}

\caption{\label{tbl-source-tab1}Source data information}

\tabularnewline

\\
\toprule
Name & URL\\
\midrule
data12 & https://archive.ics.uci.edu/dataset/518/speaker+accent+recognition\\
data13-1 & https://archive.ics.uci.edu/dataset/547/algerian+forest+fires+dataset\\
data13-2 & https://archive.ics.uci.edu/dataset/547/algerian+forest+fires+dataset\\
data20-1 & https://archive.ics.uci.edu/dataset/357/occupancy+detection\\
data20-2 & https://archive.ics.uci.edu/dataset/357/occupancy+detection\\
\addlinespace
data20-3 & https://archive.ics.uci.edu/dataset/357/occupancy+detection\\
data21 & https://archive.ics.uci.edu/dataset/537/cervical+cancer+behavior+risk\\
data22-1 & https://archive.ics.uci.edu/dataset/257/user+knowledge+modeling\\
data22-2 & https://archive.ics.uci.edu/dataset/257/user+knowledge+modeling\\
data26 & https://archive.ics.uci.edu/dataset/308/gas+sensor+array+under+flow+modulation\\
\addlinespace
data27-a11 & https://archive.ics.uci.edu/dataset/302/gesture+phase+segmentation\\
data27-a21 & https://archive.ics.uci.edu/dataset/302/gesture+phase+segmentation\\
data27-a31 & https://archive.ics.uci.edu/dataset/302/gesture+phase+segmentation\\
data27-a32 & https://archive.ics.uci.edu/dataset/302/gesture+phase+segmentation\\
data27-b11 & https://archive.ics.uci.edu/dataset/302/gesture+phase+segmentation\\
\addlinespace
data27-b12 & https://archive.ics.uci.edu/dataset/302/gesture+phase+segmentation\\
data27-b31 & https://archive.ics.uci.edu/dataset/302/gesture+phase+segmentation\\
data27-b32 & https://archive.ics.uci.edu/dataset/302/gesture+phase+segmentation\\
data27-c11 & https://archive.ics.uci.edu/dataset/302/gesture+phase+segmentation\\
data27-c12 & https://archive.ics.uci.edu/dataset/302/gesture+phase+segmentation\\
\addlinespace
data27-c31 & https://archive.ics.uci.edu/dataset/302/gesture+phase+segmentation\\
data27-c32 & https://archive.ics.uci.edu/dataset/302/gesture+phase+segmentation\\
data29 & https://archive.ics.uci.edu/dataset/282/lsvt+voice+rehabilitation\\
data32 & https://archive.ics.uci.edu/dataset/257/user+knowledge+modeling\\
data34 & https://archive.ics.uci.edu/dataset/267/banknote+authentication\\
\addlinespace
data36 & https://archive.ics.uci.edu/dataset/236/seeds\\
data37-1 & https://archive.ics.uci.edu/dataset/231/pamap2+physical+activity+monitoring\\
data37-2 & https://archive.ics.uci.edu/dataset/231/pamap2+physical+activity+monitoring\\
data37-3 & https://archive.ics.uci.edu/dataset/231/pamap2+physical+activity+monitoring\\
data37-4 & https://archive.ics.uci.edu/dataset/231/pamap2+physical+activity+monitoring\\
\addlinespace
data37-5 & https://archive.ics.uci.edu/dataset/231/pamap2+physical+activity+monitoring\\
data37-8 & https://archive.ics.uci.edu/dataset/231/pamap2+physical+activity+monitoring\\
data37-14 & https://archive.ics.uci.edu/dataset/231/pamap2+physical+activity+monitoring\\
data38 & https://archive.ics.uci.edu/dataset/230/planning+relax\\
data39-1 & https://archive.ics.uci.edu/dataset/224/gas+sensor+array+drift+dataset\\
\addlinespace
data39-2 & https://archive.ics.uci.edu/dataset/224/gas+sensor+array+drift+dataset\\
data39-3 & https://archive.ics.uci.edu/dataset/224/gas+sensor+array+drift+dataset\\
data39-4 & https://archive.ics.uci.edu/dataset/224/gas+sensor+array+drift+dataset\\
data39-5 & https://archive.ics.uci.edu/dataset/224/gas+sensor+array+drift+dataset\\
data39-6 & https://archive.ics.uci.edu/dataset/224/gas+sensor+array+drift+dataset\\
\addlinespace
data39-7 & https://archive.ics.uci.edu/dataset/224/gas+sensor+array+drift+dataset\\
data39-8 & https://archive.ics.uci.edu/dataset/225/gas+sensor+array+drift+dataset\\
data39-9 & https://archive.ics.uci.edu/dataset/226/gas+sensor+array+drift+dataset\\
data41-1 & https://archive.ics.uci.edu/dataset/194/wall+following+robot+navigation+data\\
data41-2 & https://archive.ics.uci.edu/dataset/194/wall+following+robot+navigation+data\\
\addlinespace
data41-3 & https://archive.ics.uci.edu/dataset/194/wall+following+robot+navigation+data\\
data42 & https://archive.ics.uci.edu/dataset/192/breast+tissue\\
data44 & https://archive.ics.uci.edu/dataset/181/libras+movement\\
data49-2 & https://archive.ics.uci.edu/dataset/167/hill+valley\\
data49-3 & https://archive.ics.uci.edu/dataset/168/hill+valley\\
\addlinespace
data49-4 & https://archive.ics.uci.edu/dataset/169/hill+valley\\
data50 & https://archive.ics.uci.edu/dataset/161/mammographic+mass\\
data52-1 & https://archive.ics.uci.edu/dataset/96/spectf+heart\\
data52-2 & https://archive.ics.uci.edu/dataset/96/spectf+heart\\
data53 & https://archive.ics.uci.edu/dataset/139/synthetic+control+chart+time+series\\
\addlinespace
data55-1 & https://archive.ics.uci.edu/dataset/80/optical+recognition+of+handwritten+digits\\
data55-2 & https://archive.ics.uci.edu/dataset/80/optical+recognition+of+handwritten+digits\\
data56-1 & https://archive.ics.uci.edu/dataset/81/pen+based+recognition+of+handwritten+digits\\
data56-2 & https://archive.ics.uci.edu/dataset/81/pen+based+recognition+of+handwritten+digits\\
data57 & https://archive.ics.uci.edu/dataset/39/ecoli\\
\addlinespace
data61-1 & https://archive.ics.uci.edu/dataset/54/isolet\\
data62 & https://archive.ics.uci.edu/dataset/74/musk+version+1\\
data63 & https://archive.ics.uci.edu/dataset/75/musk+version+2\\
data64-1 & https://archive.ics.uci.edu/dataset/146/statlog+landsat+satellite\\
data64-2 & https://archive.ics.uci.edu/dataset/146/statlog+landsat+satellite\\
\addlinespace
data65 & https://archive.ics.uci.edu/dataset/15/breast+cancer+wisconsin+original\\
data66 & https://archive.ics.uci.edu/dataset/62/lung+cancer\\
data67 & https://archive.ics.uci.edu/dataset/59/letter+recognition\\
data68-1 & https://archive.ics.uci.edu/dataset/50/image+segmentation\\
data68-2 & https://archive.ics.uci.edu/dataset/50/image+segmentation\\
\addlinespace
data69-1 & https://archive.ics.uci.edu/dataset/107/waveform+database+generator+version+1\\
data69-2 & https://archive.ics.uci.edu/dataset/107/waveform+database+generator+version+1\\
data70 & https://archive.ics.uci.edu/dataset/42/glass+identification\\
data71-1 & https://archive.ics.uci.edu/dataset/148/statlog+shuttle\\
data71-2 & https://archive.ics.uci.edu/dataset/148/statlog+shuttle\\
\addlinespace
data72-1 & https://archive.ics.uci.edu/dataset/149/statlog+vehicle+silhouettes\\
data72-2 & https://archive.ics.uci.edu/dataset/149/statlog+vehicle+silhouettes\\
data72-3 & https://archive.ics.uci.edu/dataset/149/statlog+vehicle+silhouettes\\
data72-4 & https://archive.ics.uci.edu/dataset/149/statlog+vehicle+silhouettes\\
data72-5 & https://archive.ics.uci.edu/dataset/149/statlog+vehicle+silhouettes\\
\addlinespace
data72-6 & https://archive.ics.uci.edu/dataset/149/statlog+vehicle+silhouettes\\
data72-7 & https://archive.ics.uci.edu/dataset/149/statlog+vehicle+silhouettes\\
data73 & https://archive.ics.uci.edu/dataset/151/connectionist+bench+sonar+mines+vs+rocks\\
data74 & https://archive.ics.uci.edu/dataset/152/connectionist+bench+vowel+recognition+deterding+data\\
NCI60 & https://github.com/natydasilva/PPforest/blob/master/data/NCI60.rda\\
\addlinespace
crab & https://github.com/natydasilva/PPforest/blob/master/data/crab.rda\\
fishcatch & https://github.com/natydasilva/PPforest/blob/master/data/fishcatch.rda\\
glass & https://github.com/natydasilva/PPforest/blob/master/data/glass.rda\\
image & https://github.com/natydasilva/PPforest/blob/master/data/image.rda\\
leukemia & https://github.com/natydasilva/PPforest/blob/master/data/leukemia.rda\\
\addlinespace
lymphoma & https://github.com/natydasilva/PPforest/blob/master/data/lymphoma.rda\\
olive & https://github.com/natydasilva/PPforest/blob/master/data/olive.rda\\
parkinson & https://github.com/natydasilva/PPforest/blob/master/data/parkinson.rda\\
wine & https://github.com/natydasilva/PPforest/blob/master/data/wine.rda\\
\bottomrule

\end{longtable}

\endgroup{}

\begin{longtable}[t]{lrrrrr}

\caption{\label{tbl-data-tab2}Overview of benchmark datasets}

\tabularnewline

\\
\toprule
dataset & Cases & Predictors & Groups & Imbalance & Correlation\\
\midrule
data12 & 329 & 12 & 6 & 0.41 & 0.38\\
data13-1 & 122 & 13 & 2 & 0.03 & 0.41\\
data13-2 & 121 & 13 & 2 & 0.29 & 0.37\\
data20-1 & 2665 & 5 & 2 & 0.27 & 0.81\\
data20-2 & 9752 & 5 & 2 & 0.58 & 0.29\\
\addlinespace
data20-3 & 8143 & 5 & 2 & 0.58 & 0.45\\
data21 & 72 & 19 & 2 & 0.42 & 0.22\\
data22-1 & 258 & 5 & 4 & 0.25 & 0.12\\
data22-2 & 145 & 5 & 4 & 0.14 & 0.18\\
data26 & 58 & 435 & 4 & 0.21 & 0.64\\
\addlinespace
data27-a11 & 1747 & 18 & 5 & 0.38 & 0.24\\
data27-a21 & 1264 & 18 & 5 & 0.35 & 0.22\\
data27-a31 & 1834 & 18 & 5 & 0.28 & 0.27\\
data27-a32 & 1830 & 32 & 5 & 0.28 & 0.13\\
data27-b11 & 1072 & 18 & 5 & 0.31 & 0.27\\
\addlinespace
data27-b12 & 1069 & 32 & 5 & 0.32 & 0.13\\
data27-b31 & 1424 & 18 & 5 & 0.21 & 0.24\\
data27-b32 & 1420 & 32 & 5 & 0.21 & 0.11\\
data27-c11 & 1111 & 18 & 5 & 0.14 & 0.27\\
data27-c12 & 1107 & 32 & 5 & 0.13 & 0.11\\
\addlinespace
data27-c31 & 1448 & 18 & 5 & 0.17 & 0.27\\
data27-c32 & 1444 & 32 & 5 & 0.17 & 0.11\\
data29 & 126 & 310 & 2 & 0.33 & 0.28\\
data32 & 403 & 6 & 4 & 0.20 & 0.11\\
data34 & 1372 & 4 & 2 & 0.11 & 0.43\\
\addlinespace
data36 & 210 & 7 & 3 & 0.00 & 0.61\\
data37-1 & 29101 & 53 & 5 & 0.23 & 0.17\\
data37-14 & 769 & 53 & 2 & 0.52 & 0.16\\
data37-2 & 14132 & 53 & 3 & 0.62 & 0.18\\
data37-3 & 11835 & 53 & 4 & 0.34 & 0.16\\
\addlinespace
data37-4 & 16357 & 53 & 5 & 0.29 & 0.15\\
data37-5 & 17560 & 53 & 5 & 0.21 & 0.16\\
data37-8 & 22990 & 53 & 9 & 0.27 & 0.17\\
data38 & 182 & 12 & 2 & 0.43 & 0.21\\
data39-1 & 445 & 128 & 6 & 0.15 & 0.55\\
\addlinespace
data39-2 & 1239 & 128 & 5 & 0.35 & 0.40\\
data39-3 & 1586 & 128 & 5 & 0.17 & 0.56\\
data39-4 & 161 & 128 & 5 & 0.32 & 0.43\\
data39-5 & 197 & 128 & 5 & 0.22 & 0.57\\
data39-6 & 2300 & 128 & 6 & 0.25 & 0.61\\
\addlinespace
data39-7 & 3613 & 128 & 6 & 0.11 & 0.68\\
data39-8 & 294 & 128 & 6 & 0.43 & 0.57\\
data39-9 & 470 & 128 & 6 & 0.10 & 0.42\\
data41-1 & 5456 & 2 & 4 & 0.34 & 0.03\\
data41-2 & 5456 & 4 & 4 & 0.34 & 0.15\\
\addlinespace
data41-3 & 5456 & 24 & 4 & 0.34 & 0.17\\
data42 & 106 & 9 & 6 & 0.08 & 0.51\\
data44 & 360 & 90 & 15 & 0.00 & 0.27\\
data49-2 & 606 & 100 & 2 & 0.03 & 1.00\\
data49-3 & 606 & 100 & 2 & 0.01 & 0.99\\
\addlinespace
data49-4 & 606 & 100 & 2 & 0.01 & 0.99\\
data50 & 830 & 5 & 2 & 0.03 & 0.23\\
data52-1 & 80 & 44 & 2 & 0.00 & 0.25\\
data52-2 & 187 & 44 & 2 & 0.84 & 0.29\\
data53 & 600 & 60 & 6 & 0.00 & 0.46\\
\addlinespace
data55-1 & 3823 & 64 & 10 & 0.00 & 0.12\\
data55-2 & 1797 & 64 & 10 & 0.01 & 0.12\\
data56-1 & 7494 & 16 & 10 & 0.01 & 0.27\\
data56-2 & 3498 & 16 & 10 & 0.01 & 0.28\\
data57 & 327 & 7 & 5 & 0.38 & 0.23\\
\addlinespace
data61-1 & 6238 & 617 & 26 & 0.00 & 0.17\\
data62 & 476 & 166 & 2 & 0.13 & 0.26\\
data63 & 6598 & 166 & 2 & 0.69 & 0.27\\
data64-1 & 4435 & 36 & 6 & 0.15 & 0.49\\
data64-2 & 2000 & 36 & 6 & 0.13 & 0.50\\
\addlinespace
data65 & 683 & 9 & 2 & 0.30 & 0.60\\
data66 & 27 & 56 & 3 & 0.07 & 0.20\\
data67 & 20000 & 16 & 26 & 0.00 & 0.18\\
data68-1 & 210 & 19 & 7 & 0.00 & 0.30\\
data68-2 & 2100 & 19 & 7 & 0.00 & 0.28\\
\addlinespace
data69-1 & 5000 & 21 & 3 & 0.01 & 0.30\\
data69-2 & 5000 & 40 & 3 & 0.01 & 0.09\\
data70 & 214 & 9 & 6 & 0.31 & 0.23\\
data71-1 & 43500 & 9 & 7 & 0.78 & 0.19\\
data71-2 & 14494 & 9 & 5 & 0.79 & 0.20\\
\addlinespace
data72-1 & 94 & 18 & 4 & 0.09 & 0.39\\
data72-2 & 94 & 18 & 4 & 0.05 & 0.40\\
data72-3 & 94 & 18 & 4 & 0.10 & 0.42\\
data72-4 & 94 & 18 & 4 & 0.12 & 0.42\\
data72-5 & 94 & 18 & 4 & 0.10 & 0.49\\
\addlinespace
data72-6 & 94 & 18 & 4 & 0.06 & 0.40\\
data72-7 & 94 & 18 & 4 & 0.06 & 0.44\\
data73 & 208 & 60 & 2 & 0.07 & 0.23\\
data74 & 990 & 12 & 11 & 0.00 & 0.22\\
NCI60 & 56 & 31 & 7 & 0.05 & 0.56\\
\addlinespace
crab & 200 & 6 & 4 & 0.00 & 0.95\\
fishcatch & 159 & 7 & 7 & 0.31 & 0.46\\
glass & 214 & 10 & 6 & 0.31 & 0.23\\
image & 2310 & 19 & 7 & 0.00 & 0.28\\
leukemia & 72 & 41 & 3 & 0.40 & 0.44\\
\addlinespace
lymphoma & 80 & 51 & 3 & 0.41 & 0.75\\
olive & 572 & 9 & 9 & 0.32 & 0.35\\
parkinson & 195 & 23 & 2 & 0.51 & 0.50\\
wine & 178 & 14 & 3 & 0.13 & 0.30\\
\bottomrule

\end{longtable}

\end{document}